\documentclass[11pt]{article}

\newlength{\figurewidth}
\usepackage{times}
\usepackage{graphicx} 
\usepackage{subfigure}
\usepackage{fullpage}

\usepackage{natbib}

\usepackage{algorithm}
\usepackage{algorithmic}

\usepackage{amsmath}
\usepackage{amssymb}
\usepackage{tikz}
\usepackage{subfigure}

\newcommand{\forspace}[1]{}

\newcommand*{\defeq}{\stackrel{\textup{def}}{=}}
\newcommand{\argmin}{\mathrm{argmin}}
\newcommand{\nez}{\text{\normalfont nez}}

\newcommand{\bb}{{\bf b}}

\newcommand{\bg}{{\bf g}}

\newcommand{\bnu}{{\boldsymbol{\nu}}}

\newcommand{\bone}{{\boldsymbol{1}}}

\newcommand{\br}{{\bf r}}

\newcommand{\bu}{{\bf u}}
\newcommand{\bv}{{\bf v}}
\newcommand{\bw}{{\bf w}}

\newcommand{\bx}{{\bf x}}

\newcommand{\bzero}{{\boldsymbol{0}}}
\newcommand{\dnote}[1]{}
\newcommand{\E}{{\bf E}}

\newcommand{\er}{\mathrm{er}}

\newcommand{\pnote}[1]{}
\newcommand{\qed}{\hfill\rule{1ex}{1.5ex} \smallskip}
\newcommand{\R}{{\bf R}}

\newcommand{\regd}{\mathrm{reg}_{D,q}}
\newcommand{\reg}{\mathrm{reg}}
\newcommand{\set}[1]{ \left\{ #1 \right\} }
\newcommand{\sign}{\mathrm{sign}}

\renewcommand{\P}{{\bf P}}
\renewcommand{\Pr}{{\bf Pr}}

\newtheorem{theorem}{Theorem}
\newtheorem{proposition}[theorem]{Proposition}
\newtheorem{lemma}[theorem]{Lemma}
\newtheorem{corollary}[theorem]{Corollary}

\newtheorem{definition}[theorem]{Definition}

\begin{document}

\title{On the Inductive Bias of Dropout}

\author{David P.\ Helmbold \\
        UC Santa Cruz \\
        dph@soe.ucsc.edu
       \and
       Philip M.\ Long \\
       Microsoft \\
       plong@microsoft.com
}

\maketitle

\begin{abstract}
Dropout is a simple but effective technique for learning in neural networks and other settings.
A
sound  theoretical understanding of dropout is needed to determine when dropout should be applied and how to use it most effectively.
In this paper we continue the exploration of dropout as a regularizer pioneered by Wager, et.al.
We focus on linear classification where a convex proxy to the misclassification loss (i.e. the logistic loss used in logistic regression) is minimized.
We show:
\begin{itemize}
\item when the dropout-regularized criterion has a unique minimizer,
\item when the dropout-regularization penalty goes to infinity with the weights, and when it remains bounded,
\item that the dropout regularization can be non-monotonic as individual weights increase from 0, and
\item that the dropout regularization penalty may \emph{not} be convex.
\end{itemize}
This last point is particularly surprising because
the combination of dropout regularization with any convex loss proxy is always a convex function.

In order to contrast dropout regularization with $L_2$ regularization, we formalize the
notion of when different sources are more compatible with different regularizers.
We then exhibit distributions that are provably more compatible with dropout regularization than $L_2$ regularization, and vice versa.
These sources provide additional insight into how the inductive biases of dropout and $L_2$ regularization differ.  We provide some similar results for
$L_1$ regularization.
\end{abstract}


\newpage
\section{Introduction}

%
Since its prominent role in a win of the ImageNet Large Scale Visual Recognition Challenge
\citep{Hin12,HSKSS12}, there has been intense interest in dropout
(see the work by \citet{Dah12,DenEtAl13,DSH13,WanEtAl13,WWL13,BS13,VKW14}).
This paper studies the inductive bias of dropout:
when one chooses to train with dropout, what prior preference over models
results?  
We show that dropout training shapes the learner's search space in a much
different way than $L_1$ or  $L_2$ regularization. 
Our results shed new insight into why dropout prefers rare features,
how the dropout probability affects the strength of regularization,
and how dropout restricts the co-adaptation of weights.

Our theoretical study will concern learning  a linear classifier
via convex optimization.  The learner wishes to find a parameter vector
$\bw$ so that, for a random feature-label pair
$(\bx,y) \in \R^n \times \{ -1,1 \}$ drawn from some joint
distribution $P$, the probability that $\sign(\bw \cdot \bx) \neq y$
is small.
It does this by using training data
to try to minimize $\E(\ell(y \bw \cdot \bx))$, where
$\ell(z) = \ln (1 + \exp(-z))$ is the loss function
associated with
logistic regression.

We have chosen to focus on this
problem for several reasons.
First, the inductive bias of dropout is not well understood even in this simple setting.
Second, linear classifiers
remain a popular choice for practical problems, especially in the
case of very high-dimensional data.
Third, we view a thorough understanding of dropout in this setting
as a mandatory prerequisite to
understanding the inductive bias of dropout when
applied in a deep learning architecture.
This is especially true when the preference
over deep learning models is decomposed
into preferences at each node.
In any case,
the setting that we are studying faithfully describes the inductive
bias of a deep learning system at its output nodes.

We will borrow the following clean and illuminating description of
dropout as artificial noise due to \citet{WWL13}.  An algorithm for
linear classification using loss $\ell$
and dropout updates its parameter vector $\bw$ online, using
stochastic gradient descent.  Given an example $(\bx,y)$, the dropout
algorithm independently perturbs each feature $i$ of
$\bx$: with probability $q$, $x_i$ is replaced with $0$, and, with
probability $p = 1-q$, $x_i$ is replaced with $x_i/p$.
Equivalently, $\bx$ is replaced by $\bx + \bnu$, where
\[
\nu_i = \left\{ \begin{array}{ll}
          -x_i & \mbox{with probability $q$} \\
          (1/p - 1) x_i & \mbox{with probability $p = 1-q$}
                \end{array}
                            \right.
\]
before performing the stochastic gradient update step.
(Note that, while $\bnu$ obviously depends on $\bx$,
if we sample the components of $\bb \in \{-1, 1/p - 1 \}^n$
independently of one another and $\bx$,
by choosing $b_i = -1$ with
the dropout probability $q$,
then we may write $\nu_i = b_i x_i$.)

Stochastic gradient descent is known to converge under a broad variety
of conditions \citep{KY97}.  Thus, if we abstract away sampling
issues as done by
\citet{Bre04,Zha04convex,BJM06,LS10},
we are led to
consider
\[
\bw^* \defeq \argmin_{\bw} \E_{(\bx,y) \sim P, \bnu}
                  ( \ell(y \bw \cdot (\bx + \bnu)))
\]
as dropout can be viewed as a stochastic gradient update of this
global objective function.
We call this objective the {\em dropout criterion},  and it can be viewed as a risk
on the dropout-induced distribution.
(Abstracting away sampling issues is
consistent with our goal of concentrating 
on the inductive bias of the algorithm.
From the point of view of a bias-variance decomposition, we do not intend
to focus on the large-sample-size case, where the variance is small, but
rather to focus on the contribution from the bias 
where $P$ could be an empirical sample distribution.
)

We start with the observation of
\citet{WWL13} that the dropout criterion may be decomposed as
\begin{equation}
  \label{e:dropout.decompose}
\E_{(\bx,y) \sim P, \bnu}
     ( \ell(y \bw \cdot (\bx + \bnu)))
 = \E_{(\bx,y) \sim P}
     ( \ell(y \bw \cdot \bx))
 + \regd(\bw),
\end{equation}
where $\regd$ is non-negative, and depends only on the marginal
distribution $D$ over the feature vectors $\bx$ (along with the
dropout probability $q$), and not on the
labels.  This leads naturally to a view of dropout as a regularizer.

A popular style of learning algorithm minimizes an objective function
like the RHS of (\ref{e:dropout.decompose}), but where
$\regd(\bw)$ is replaced by a norm of $\bw$.  One motivation for
algorithms in this family is to first replace the training error
with a convex proxy to make optimization tractable, and then to regularize
using a convex penalty such as a norm, so that the objective function remains
convex.

We show that $\regd(\bw)$ formalizes a preference for classifiers
that assign a very large weight to a single feature.  This preference
is stronger than what one gets from a penalty proportional to
$|| \bw ||_1$.  In fact, we show that, despite the convexity of the
dropout risk, $\regd$ is {\em not} convex, so that dropout
provides a way to realize the inductive bias arising from a non-convex
penalty, while still enjoying the benefit of convexity in the
overall objective function (see the plots in Figures~\ref{f:nonmonotone}, \ref{f:L2WinsPlots} and~\ref{f:drop.wins.plots}).
Figure~\ref{f:nonmonotone} shows the even more surprising result that the dropout regularization penalty
is not even monotonic in the absolute values of the individual weights.

It is not hard to see
that $\regd({\bf 0}) = 0$.  Thus, if $\regd({\bf w})$ is greater than
the expected loss incurred by ${\bf 0}$ (which is $\ln 2$), then it might
as well be infinity, because dropout will prefer ${\bf 0}$ to $\bw$.  However,
in some cases, dropout never reaches this extreme -- it remains willing
to use a model, even if its parameter is very large, unlike methods
that use a convex penalty.  In particular,
\[
\regd(w_1,0,0,0,...,0) < \ln 2
\]
for all $D$,
no matter how large $w_1$ gets; of course, the same is true
for the other features.  On the other hand, except for some
special cases (which are detailed in the body of the paper),
\[
\regd(c w_1,c w_2,0,0,...,0)
\]
goes to infinity with $c$.  It follows that $\regd$ cannot
be approximated to within any factor, constant or otherwise, by a
convex function of $\bw$.

To get a sense of which sources dropout can be
successfully applied to, we compare dropout with an algorithm that
regularizes using $L_2$, by minimizing the \emph{$L_2$ criterion}:
\begin{equation}
\label{e:elltwo.objective}
\E_{(\bx,y) \sim P}
     ( \ell(y \bw \cdot \bx))
    + \frac{\lambda}{2} || \bw ||_2^2.
\end{equation}
Will will use ``$L_2$'' as a shorthand to refer to an algorithm
that minimizes (\ref{e:elltwo.objective}).  Note that $q$, the
probability of dropping out an input feature, plays a role in dropout
analogous to $\lambda$.
In particular, as $q$ goes to zero the examples remain unperturbed and
the dropout regularization has no effect.

Informally, we say that joint probability
distributions $P$ and $Q$ {\em separate} dropout from $L_2$ if,
when the same parameters $\lambda$ and $q$ are used for both $P$ and
$Q$, then using dropout leads to a much more accurate hypothesis for
$P$, and using $L_2$ leads to a much more accurate hypothesis for
$Q$.
This enables us to illustrate the inductive biases of the
algorithms through the use of contrasting sources
that either align or are incompatible with the algorithms' inductive
bias.
Comparing with another regularizer helps to restrict these
illustrative examples to ``reasonable'' sources, which can be handled
using another regularizer.  Ensuring that the same values of the
regularization parameter are used for both $P$ and $Q$ controls for
the amount of regularization, and ensures that the difference is due to
the model preferences of the respective regularizers.
This style
of analysis is new, as far as we know, and may be a useful tool for studying the inductive biases of
other algorithms and in other settings.

{\bf Related previous work.}  Our research builds on the work of
\citet{WWL13}, who analyzed dropout  for random
$(x,y)$ pairs where the distribution of $y$ given $x$ comes from a member
of the exponential family, and the quality of a model is evaluated using
the log-loss.  They pointed out that, in these cases, the
dropout criterion can be decomposed into the original loss and a
term that does not depend on $y$, which therefore can be viewed as a
regularizer.  They then proposed an approximation to this dropout
regularizer, discussed its relationship with other regularizers and
training algorithms, and evaluated it experimentally.  \citet{BS13}
exposed properties of dropout when viewed as an ensemble method (see
also \citet{BAP14}).  \citet{VKW14} showed that applying dropout
for online learning in the experts setting leads to algorithms that
adapt to important properties of the input without requiring doubling
or other parameter-tuning techniques, and
\citet{AbeEtAl14} analyzed a class of methods including dropout
by viewing these methods as smoothers.  The impact of dropout on
generalization (roughly, how much dropout restricts the search space
of the learner, or, from a bias-variance point of view, its impact on
variance) was studied by \citet{WanEtAl13} and \citet{WFWL14}. The
latter paper
considers a variant of dropout compatible with a poisson source, and shows that under some assumptions
this dropout variant converges more quickly to its infinite sample limit
than non-dropout training,
and that the Bayes-optimal predictions are preserved under the modified dropout distribution.
Our results complement theirs by focusing on the effect of the original dropout on the algorithm's bias.

%

Section~\ref{s:prelim} defines our notation and characterizes when the dropout criterion has a unique minimizer.
Section~\ref{s:dropout.regularizer} presents many additional properties of  the dropout regularizer.
Section~\ref{s:separation.def} formally defines when two distributions separate two algorithms or regularizers.
Sections~\ref{s:2d.L2.wins} and~\ref{s:2d.drop.wins}  give sources over $\R^2$ that 
separate dropout and $L_2$.  Section~\ref{s:L1} provides plots demonstrating that the same distributions separated dropout from $L_1$ regularization.
Sections~\ref{s:L2.manyfeatures} and~\ref{s:dropout.wins.large.n}
give separation results from $L_2$ with many features.


\section{Preliminaries}
\label{s:prelim}

We use $\bw^*$ for the optimizer of the dropout criterion, $q$ for the probability that a feature is dropped out, and $p=1-q$ for the probability that a feature is kept  throughout the paper.
As in the introduction, if $X \subseteq \R^n$ and $P$ is a joint distribution over
$X \times\{-1, 1\}$, define
\begin{equation}
\label{e:dropout.optimizer}
\bw^*(P,q) \defeq \argmin_{\bw} \E_{(\bx,y) \sim P, \bnu}
                  ( \ell(y \bw \cdot (\bx + \bnu)))
\end{equation}
where $\nu_i = b_i x_i$ for $b_1,...,b_n$ sampled independently at
random from $\{-1, 1/p - 1\}$ with $\Pr(b_i = 1/p - 1) = p = 1-q$,
and $\ell(z)$ is the logistic loss function:
\[
\ell(z) = \ln (1 + \exp(-z)).
\]

For some analyses, an alternative
representation of $\bw^*(P,q)$ will be easier to work with.
Let
$r_1,...,r_n$ be sampled randomly from $\{ 0, 1 \}$, independently of
$(\bx, y)$ and one another, with $\Pr(r_i = 1) = p$.
Defining $\br \odot \bx = (x_1 r_1,..., x_n r_n)$,
we have the equivalent definition
\begin{equation}
\label{e:r.odot}
\bw^*(P,q) = p \; \argmin_{\bw} \E_{(\bx,y) \sim P, \br}
                  ( \ell(y \bw \cdot (\br  \odot \bx))).
\end{equation}
To see that they are equivalent, note that
\begin{align*}
  \E( \ell(y \bw \cdot (\bx + \bnu)))
  & = \E\left( \ell\left(y \bw \cdot \left(\frac{\br  \odot \bx}{p}
             \right) \right)  \right) \\
  &           = \E ( \ell(y (\bw/p) \cdot (\br  \odot \bx))).
\end{align*}
Although this paper focuses on the logistic loss, the above definitions can be used for any loss function $\ell()$.
Since the dropout criterion is an expectation of $\ell()$, we have the following obvious consequence.
\begin{proposition}
%
If loss $\ell(\cdot)$ is convex, then the dropout criterion is also a convex function of $\bw$.
\end{proposition}

Similarly, we use $\bv$ for the optimizer of the $L_2$ regularized criterion:
\begin{equation}
\label{e:L2.optimizer}
\bv(P,\lambda) \defeq \argmin_{\bw} \E_{(\bx,y) \sim P}
                    ( \ell(y \bw \cdot \bx) ) + \frac{\lambda}{2} || \bw ||^2.
\end{equation}

It is not hard to see that the
$\frac{\lambda}{2} || \bw ||^2$ term implies that
$\bv(P,\lambda)$ is always well-defined.  On the other hand,
$\bw^*(P,q)$ is {\em not} always well-defined, as can be seen by
considering any distribution concentrated on a single example.
This motivates the following definition.

%
%
%
%
%
%
%
%

\begin{definition}
Let $P$ be a joint distribution with support contained in $\R^n \times
\{-1,1\}$.
A feature $i$ is \underline{\em perfect modulo ties} for $P$ if either
$y x_i  \geq 0$ for all $\bx$ in the support of $P$,  or
$y x_i \leq 0$ for all $\bx$ in the support of $P$.
\end{definition}
Put another way, $i$ is perfect modulo ties if there is a linear classifier
that only pays attention to feature $i$ and is perfect on the part of $P$
where $x_i$ is nonzero.

\begin{proposition}
For all finite domains $X \subseteq \R^n$, all distributions
$P$ with support in $X$, and all $q \in (0,1)$, we have that
$\E_{(\bx,y) \sim P, \br} ( \ell(y \bw \cdot (\br \odot \bx)))$
has a unique minimum in $\R^n$  if and only if no feature is perfect
modulo ties for $P$.\end{proposition}
{\bf Proof:}  Assume for contradiction that feature $i$ is perfect modulo ties
for $P$
and some $\bw^{\circledast}$ is the unique minimizer of
$\E_{(\bx,y) \sim P, \br} ( \ell(y \bw \cdot (\br \odot \bx)))$.
Assume w.l.o.g.\ that
$y x_i \geq 0$ for all $\bx$ in the support of $P$
(the case where $y x_i \leq 0$ is analogous).
Increasing $w^\circledast_i$ keeps the loss unchanged
on examples where $x_i=0$ and decreases the
loss on the other examples in the support of $P$, contradicting the assumption that $\bw^\circledast$ was a unique minimizer of the expected loss.

Now, suppose then each feature $i$ has both examples where $y x_i >0$ and examples where $y x_i < 0$ in the support of $P$.
Since the support
of $P$ is finite, there is a positive lower bound on the probability of any example in the support.
With probability
$p (1-p)^{n-1}$, component $r_i$ of random vector $\br$ is non-zero and the remaining $n-1$ components are all zero.
Therefore as $w_i$ increases without bound in the positive or negative direction,
$\E_{(\bx,y) \sim P, \br} ( \ell(y \bw \cdot (\br \odot \bx)))$ also increases
without bound.
Since
$\E_{(\bx,y) \sim P, \br} ( \ell(y {\bf 0} \cdot (\br \odot \bx))) = \ln 2$,
there is a value $M$ depending only on distribution $P$ and the dropout probability such that
minimizing $\E_{(\bx,y) \sim P, \br} ( \ell(y {\bf w} \cdot (\br \odot \bx)))$
over $\bw \in [-M,M]^n$ is equivalent to minimizing
$\E_{(\bx,y) \sim P, \br} ( \ell(y {\bf w} \cdot (\br \odot \bx)))$ over
$\R^n$.  Since $\Pr_{(\bx,y)} ( x_i = 0) \neq 1$ for all $i$,
$\{ \br \odot \bx: \br \in \{ 0, 1 \}^n, \bx \in X \}$ has full rank
and therefore
$\E_{(\bx,y) \sim P, \br} ( \ell(y {\bf w} \cdot (\br \odot \bx)))$ is strictly
convex.  Since a strictly convex function defined on a
compact set has a unique minimum, $\E_{(\bx,y) \sim P, \br} ( \ell(y {\bf w} \cdot (\br \odot \bx)))$
has a unique minimum on $[-M,M]^n$, and therefore on $\R^n$. \qed

See Table~\ref{t:notation} for a summary of the notation used in the paper.

\begin{table}
\centerline{
\begin{tabular}{| c | l |}
\hline
$\bx=(x_1, \ldots, x_n)$ & feature vector in $\R^n$ 		\\ \hline
$y$					& label in $\{-1,+1\}$ 		\\ \hline
$\bw = (w_1, \ldots, w_n)$ & weight vector in $\R^n$ 	\\ \hline
$\ell(y \bw\cdot \bx)$		& loss function, generally the logistic loss: $\ln(1+\exp(-y \bw \cdot \bx))$ \\ \hline
$P$, $Q$				& source distributions over $(\bx,y)$ pairs, varies by section \\ \hline
$D$					& marginal distribution over $\bx$ \\ \hline
$q$					& feature dropout probability in $(0,1)$ \\ \hline
$p=1-q$				& probability of keeping a feature \\ \hline
$\lambda$			& $L_2$ regularization parameter \\ \hline
$\bnu = (\nu_1, \ldots, \nu_n)$ & additive dropout noise, $\nu_i \in \{-x_i, x_i/p -x_i\}$  \\ \hline
$\br = (r_1, \ldots, r_n)$ & multiplicative dropout noise, $r_i \in \{0,1\}$  \\ \hline
$\odot$				& component-wise product: $\br \odot \bx = (r_1 x_1, \ldots, r_n x_n)$ \\ \hline
$\bw^*(P,q)$ and $\bw^*$ 	& minimizer of dropout criterion: $\E( \ell(y \;\bw \cdot (\bx + \bnu)))$ \\ \hline
$\bw^\circledast = \bw^*/p$ & minimizer of expected loss $\E(\ell (y \; \bw \cdot (\br \odot \bx) ) )$ \\ \hline
$\bv(P,\lambda)$ and $\bv$	& minimizer of $L_2$-regularized loss \\ \hline
$\regd(\bw)$			& regularization due to dropout \\ \hline
$J$, $K$				& criteria to be optimized, varies by sub-section \\ \hline
$g(\bw)$, $\bg$		& gradients of the current criterion \\ \hline
$\er_P(\bw)$			& 0-1 classification generalization error of $\sign(\bw \cdot x)$ \\ \hline
\end{tabular}
}
\caption{Summary of notation used throughout the paper.}
\label{t:notation}
\end{table}



\section{Properties of the Dropout Regularizer}
\label{s:dropout.regularizer}


We start by rederiving the regularization function corresponding to dropout training previously presented in \cite{WWL13}, specialized to our context and using
our notation.
The first step is to write $\ell(y \bw \cdot \bx)$ in an alternative way that
exposes some symmetries:
{\small
\begin{align}
\nonumber
 \ell(y \bw \cdot \bx)
&  =  \ln (1 + \exp(-y \bw \cdot \bx)) \\
\nonumber
&  =  \ln \left(\frac{\exp(y (\bw \cdot \bx)/2) + \exp(-y(\bw \cdot \bx)/2)}
                   {\exp(y (\bw \cdot \bx)/2)} \right) \\
\label{e:ell.logistic}
&  =  \ln \left(\frac{\exp((\bw \cdot \bx)/2) + \exp(-(\bw \cdot \bx)/2)}
      {\exp(y (\bw \cdot \bx)/2)} \right).
\end{align}
}
This then implies
{\small
\begin{align*}
\regd(\bw) & = \E(\ell(y \bw \cdot (\bx + \bnu))) - \E(\ell(y \bw \cdot \bx)) \\
& = \E \left(
  \ln \left(
    \frac{\exp((\bw \cdot (\bx + \bnu))/2) + \exp(-(\bw \cdot (\bx + \bnu))/2)}
         {\exp(y (\bw \cdot (\bx + \bnu))/2)} 
             \times
         \frac{\exp(y (\bw \cdot \bx)/2)}
              {\exp((\bw \cdot \bx)/2) + \exp(-(\bw \cdot \bx)/2)}
              \right) \right) \\
&= \E \Big(
  \ln \left(
    \frac{\exp((\bw \cdot (\bx \!+\! \bnu))/2) \!+\! \exp(-(\bw \cdot (\bx \!+\! \bnu))/2)}
     {\exp((\bw \cdot \bx)/2) \!+\! \exp(-(\bw \cdot \bx)/2)}
     \right) 
   - y (\bw \cdot \bnu)/2 \Big).
\end{align*}
}
Since $\E(\bnu) = {\bf 0}$, we get the following.
\begin{proposition} \citep{WWL13}
{\small
\begin{equation}
  \label{e:dropout.reg}
\regd(\bw)   = \E \left(
  \ln \left(
    \frac{\exp(\bw \cdot (\bx + \bnu)/2) + \exp(-\bw \cdot (\bx + \bnu)/2)}
     {\exp((\bw \cdot \bx)/2) + \exp(-(\bw \cdot \bx)/2)}
     \right) \right) .
\end{equation}
}
\end{proposition}

\begin{figure}
  \begin{center}
\begin{tabular}{c c}
\includegraphics[width=\figurewidth]{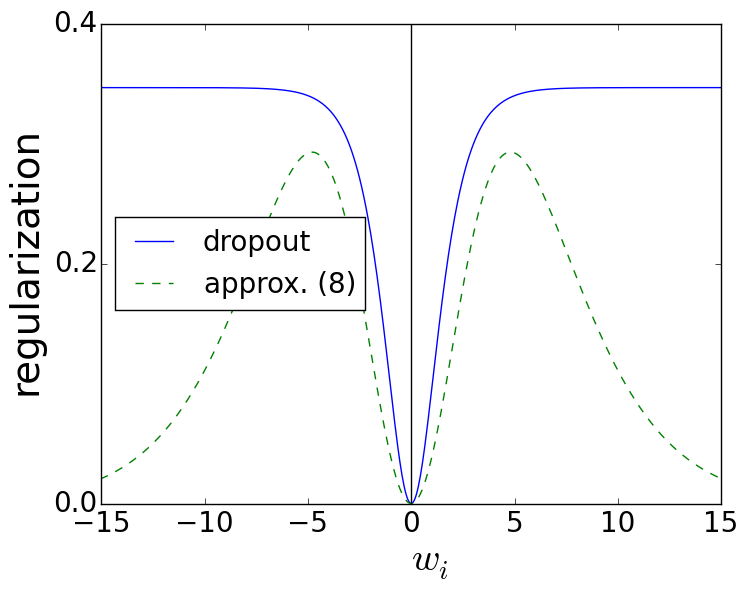} &
\includegraphics[width=\figurewidth]{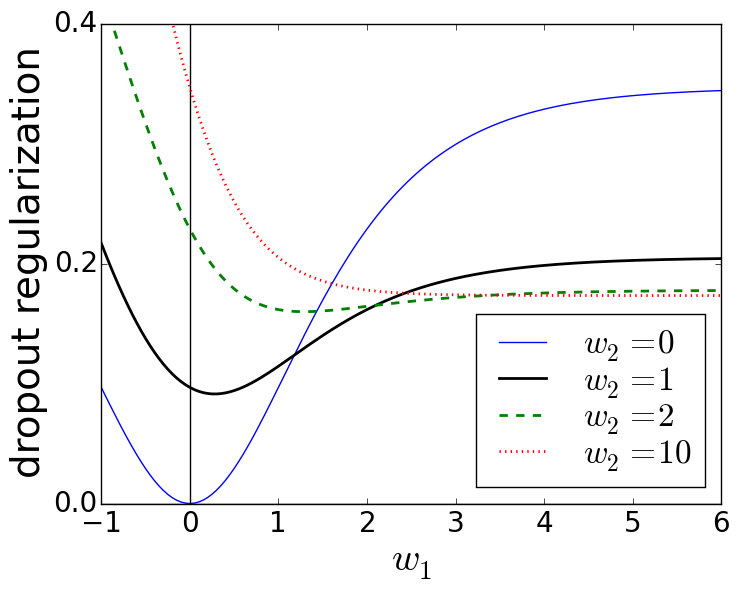} \\
\end{tabular} \\
  \end{center}
\caption{The $p=1/2$ dropout regularization for $\bx=(1,1)$ as a function of $w_i$ when the other weights are 0 together with its approximation~(\ref{e:WWL13}) (left)
and as a function of $w_1$
for different values of the second weight (right).}
\label{f:nonmonotone}
\end{figure}

Using a Taylor expansion,
\citet{WWL13} arrived at the following approximation:
\begin{equation}
  \label{e:WWL13}
\frac{q}{2 (1-q)} \sum_i w_i^2
    \E_{\bx} \left(\frac{x_i^2}{(1 + \exp(- \frac{\bw \cdot \bx}{2}))(1 + \exp( \frac{\bw \cdot \bx}{2})} \right).
\end{equation}
This approximation suggests two properties:
the strength of  the regularization penalty decreases exponentially in the prediction confidence $| \bw \cdot \bx |$,
and that the regularization penalty goes to infinity as the dropout probability $q$ goes to 1.
However, $\bw \cdot \bnu$ can be quite large,
making a second-order Taylor expansion inaccurate.%
\footnote{\citet{WWL13} experimentally evaluated the accuracy of a
  related approximation in the case that, instead of using dropout,
  $\bnu$ was distributed according to a zero-mean gaussian.  }
In fact, the analysis in this section suggests that the regularization penalty does not decrease with the confidence 
and that the regularization penalty increases linearly with $q=1-p$
(Figure~\ref{f:nonmonotone}, Theorem~\ref{t:bounded.regularization}, Proposition~\ref{p:limit.equals}).

The following propositions show that $\regd(\bw)$ satisfies at least some of the intuitive properties of a 
regularizer.

\begin{proposition}
$\regd(\bzero)=0$.
\end{proposition}

\begin{proposition} \citep{WWL13}
\label{p:non-neg}
The contribution of each $\bx$ to the regularization penalty (\ref{e:dropout.reg}) is non-negative:
for all $\bx$,
{\small
\[
 \E_{\bnu}  \left(
  \ln \left(
    \frac{\exp((\bw \cdot (\bx + \bnu))/2) + \exp(-(\bw \cdot (\bx + \bnu))/2)}
     {\exp((\bw \cdot \bx)/2) + \exp(-(\bw \cdot \bx)/2)}
     \right) \right) 
     \geq 0.
     \]
     }
\end{proposition}
{\bf Proof:}  The proposition follows from Jensen's Inequality. \qed

The $\bw^*(P,q)$ vector learned by dropout training
minimizes $\E_{(\bx,y) \sim P}  (\ell(y \bw \cdot \bx)) + \regd(\bw)$.
However, the $\bzero$  vector has $\ell(y \bzero \cdot \bx) = \ln (2)$  and $\regd(\bzero)=0$, implying:
\begin{proposition}
$\regd(\bw^*) \leq \ln (2)$.
\end{proposition}
Thus any regularization penalty greater than $\ln(2)$ is effectively equivalent to a regularization penalty of $\infty$.

We now present new results based on analyzing the exact $\regd(\bw)$.
The next properties show that the dropout regularizer is emphatically  \emph{not} like other convex or norm-based regularization penalties
in that the dropout regularization penalty always remains bounded when a single component of the weight vector goes to infinity
(see also Figure~\ref{f:nonmonotone}).

\begin{theorem}
\label{t:bounded.regularization}
For all dropout probabilities $1-p \in (0,1)$, all $n$, all marginal distributions $D$ over $n$-feature vectors,
and all indices $1\leq i \leq n$,
\[
  \sup_{w_i} \regd( \underbrace{0, \ldots, 0}_{i-1}, w_{i},  \underbrace{0, \ldots , 0}_{n-i}) 
  \leq \Pr_D(x_i \neq 0) (1-p) \ln (2) \;<\; \ln 2.
  \]
\end{theorem}
\textbf{Proof:}
Fix arbitrary $n$, $p$, $i$, and $D$.
We have
{\small
\begin{align*}
& \regd ( \underbrace{0, \ldots, 0}_{i-1}, w_{i},  \underbrace{0, \ldots , 0}_{n-i}) \\
&=
\E_{\bx,\bnu} \left(
	 \ln \left( \frac{\exp(-w_i (x_i \!+\! \nu_i )/2)\!+\!\exp(w_i (x_i \!+\! \nu_i)/2)}{\exp(- w_i x_i / 2)\!+\!\exp(w_i x_i / 2)} \right)
	 \right).
\end{align*}
}
Fix an arbitrary $\bx$ in the support of $D$ and examine the expectation over $\bnu$ for that $\bx$.
Recall that $x_i + \nu_i$ is 0 with probability $1-p$ and is $x_i/p$ with probability $p$, and
we will use the substitution $z = |w_i x_i| / 2$.
\begin{align}
\E_{\bnu} & \left(
	 \ln \left( \frac{\exp(\frac{-w_i (x_i + \nu_i )}{2})+\exp(\frac{w_i (x_i + \nu_i)}{2})}{\exp(\frac{- w_i x_i}{2})+\exp(\frac{w_i x_i}{2})} \right)
	 \right)  \label{e:expect.bnu} \\
        &= (1-p) \ln(2) + p \ln \left( \exp( \frac{z}{p}) + \exp(\frac{- z}{p}) \right) - \ln \left( \exp(z) + \exp( -z ) \right).	 \label{e:zform}
\end{align}
We now consider cases based on whether or not $z$ is 0.
When $z = 0$ (so either $w_i$ or $x_i$ is $0$) then (\ref{e:zform}) is also 0.

If $z \neq 0$ then consider the derivative of~(\ref{e:zform}) w.r.t.~$z$, which is
\[
\frac{ \exp(z/p) - \exp(-z / p)}{ \exp(z/p) + \exp(-z/p) } - \frac{ \exp(z) - \exp(-z)}{ \exp(z) + \exp(-z)}.
\]
This derivative is positive since $z > 0$ and $0< p < 1$.
Therefore~(\ref{e:zform}) is  bounded by its limit as $z \rightarrow \infty$, which is $(1-p) \ln (2)$, in this case.

Since~(\ref{e:expect.bnu}) is  0 when $x_i =0$ and is bounded by $(1-p) \ln (2)$ otherwise,
the expectation over $\bx$ of (\ref{e:expect.bnu}) is bounded $\Pr_D(x_i \neq 0) (1-p) \ln (2)$, completing the proof.
\qed

Since line~(\ref{e:zform})  is derived using a chain of equalities,
the same proof ideas can be used to show that Theorem~\ref{t:bounded.regularization} is tight.
\begin{proposition}
\label{p:limit.equals}
Under the conditions of Theorem~\ref{t:bounded.regularization},
  \[
 \lim_{w_i \rightarrow \infty} \regd( \underbrace{0, \ldots, 0}_{i-1}, w_{i},  \underbrace{0, \ldots , 0}_{n-i})
 = \Pr_D(x_i \neq 0) (1-p) \ln (2).
 \]
\end{proposition}

Note that this bound on the regularization penalty depends neither on the range nor expectation of  $x_i$.
In particular, it has a far different character than the
approximation of Equation~(\ref{e:WWL13}).

In Theorem~\ref{t:bounded.regularization} the other weights are fixed at 0 as $w_i$ goes to infinity.
An additional assumption implies that the regularization penalty remains bounded even when the other components are non-zero.
Let $\bw$ be a weight vector such that for all $\bx$ in the support of $D$ and dropout noise vectors $\bnu$ we have $| \sum_{j \neq i} w_j(x_j+\nu_j) | \leq M$
for some bound $M$ (this implies that $| \sum_{j \neq i} w_j x_j | \leq M$ also).
Then
\begin{align}
  \nonumber
 \regd(\bw) &= \E_{\bx, \bnu}  \left( \left(
			\frac{\exp(\frac{\bw \cdot (\bx \!+\! \bnu)}{2}) \!+\! \exp(-\frac{\bw \cdot (\bx \!+\! \bnu)}{2})}
			     {\exp(\frac{\bw \cdot \bx}{2}) \!+\! \exp(-\frac{\bw \cdot \bx}{2})}  \right) \right)   \nonumber \\
                             &\leq  \E_{x_i, \nu_i} \left( \log \left(  \frac{\exp(\frac{M - w_i (x_i \!+\! \nu_i)}{2}
                             \!+\! \exp(\frac{M + w_i (x_i \!+\! \nu_i)}{2})}{\exp(-\frac{M - w_i x_i}{2} \!+\! \exp(-\frac{M + w_i x_i}{2})}  \right) \right)   \nonumber \\
\label{e:M.bound}
&\leq M \!+\! \E_{x_i, \nu_i} \left( \log \left(  \frac{\exp(-\frac{w_i x_i \!+\! \nu_i}{2}) \!+\! \exp(\frac{w_i (x_i \!+\! \nu_i)}{2})}{\exp(\frac{-w_i x_i}{2}) \!+\! \exp(\frac{w_i x_i}{2})}   \right) \right) .
\end{align}
Using (\ref{e:M.bound}) instead of the first line in Theorem~\ref{t:bounded.regularization}'s proof gives the following.
\begin{proposition}
\label{p:bounded.reg.arb.w}
Under the conditions of Theorem~\ref{t:bounded.regularization},
if the weight vector $\bw$ has the property that $| \sum_{j \neq i} w_j(x_j +\nu_j ) | \leq M$ for each $\bx$ in the support of $D$ and
all of its corresponding dropout noise vectors $\bnu$ then
\[
 \sup_{\omega} \regd(w_1, w_2, \ldots, w_{i-1}, \omega,  w_{i+1}, \ldots ,w_n) 
 \leq M +  \Pr_D(x_i \neq 0) (1-p) \ln (2).
 \]
\end{proposition}
Proposition~\ref{p:bounded.reg.arb.w} shows that the regularization penalty starting from a non-zero initial weight vector remains bounded as any one of its
components goes to infinity.   On the other hand,  unless $M$ is small, the bound 
will be larger than the dropout criterion for the zero vector.
This is a natural consequence as the starting weight vector $\bw$ could already have a large regularization penalty.


The derivative of~(\ref{e:zform}) in the proof of Theorem~\ref{t:bounded.regularization} implies that the dropout regularization penalty is monotonic in $|w_i|$
when the other weights are zero.
Surprisingly, this is does \emph{not} hold in general.
The dropout regularization penalty due to a single example (as in Proposition~\ref{p:non-neg})
can be written as
\[
 \E_{\bnu} \left(
  \ln \left( \textstyle \exp(  \frac{\bw \cdot (\bx + \bnu)}{2}) + \exp(\frac{-\bw \cdot (\bx + \bnu)}{2}) \right)  \right)
  - \ln \left(  \textstyle \exp( \frac{\bw \cdot \bx}{2}) + \exp(\frac{-\bw \cdot \bx}{2}) \right)  .
  \]
Therefore if increasing a weight makes the second logarithm increase faster than the expectation of the first,
then the regularization penalty decreases even as the weight increases.
This happens when the $w_i x_i$ products tend to have the same sign.
The regularization penalty as a function of $w_1$ for the single example
$\bx = (1,1)$, $p=1/2$, and $w_2$ set to various values is plotted in Figure~\ref{f:nonmonotone}%
\footnote{Setting $\bx=(1,1)$ is in some sense without loss of generality as
the prediction and dropout regularization values for any $\bw$, $\bx$ pair are identical to the
values for $\tilde \bw$, $\bone$ when each $\tilde w_i = w_i x_i$.}
.
This gives us the following.
\begin{proposition}
Unlike p-norm regularizers,
the dropout regularization penalty~$\regd(\bw)$ is \underline{not} always monotonic in the individual weights.
\end{proposition}
In fact, the dropout regularization penalty can decrease as weights move up from 0.
\begin{proposition}
  \label{p:decrease.from.0}
Fix $p = 1/2$, $w_2>0$, and an arbitrary $\bx \in (0,\infty)^2$.  Let $D$ be the distribution concentrated on $\bx$.
Then $\regd(w_1,w_2)$ locally \underline{decreases} as $w_1$ \underline{increases} from $0$.
\end{proposition}
Proposition~\ref{p:decrease.from.0} is proved in Appendix~\ref{a:decrease.from.0}.


We now turn to the dropout regularization's behavior when two weights vary together.
If any features are always zero then their weights can go to $\pm \infty$ without affecting either the predictions or $\regd(\bw)$.
Two linearly dependent features might as well be one feature.
After ruling out degeneracies like
these, we arrive at the following theorem, which is proved in Appendix~\ref{a:two.unbounded}.

\begin{theorem}
\label{t:two.unbounded}
Fix an arbitrary distribution $D$ with support in $\R^2$, weight vector $\bw \in \R^2$, and non-dropout probability $p$.
If there is an $\bx$ with positive probability
under $D$ such that $w_1 x_1 $ and $w_2 x_2$ are both non-zero and have different signs,
then the regularization penalty $\regd(\omega \bw)$ goes to infinity as $\omega$ goes to $\pm \infty$.
\end{theorem}

The theorem can be straightforwardly generalized to the case $n > 2$; except
in degenerate cases, sending two weights to infinity together will
lead to a regularization penalty approaching infinity.

Theorem~\ref{t:two.unbounded} immediately leads to the following corollary.
\begin{corollary}
For a distribution $D$ with support in $\R^2$,
if there is an $\bx$ with positive probability
under $D$ such that $x_1 \neq 0$ and $x_2 \neq 0$, then
there is a $\bw$ such that for any $q \in (0,1)$,
the regularization penalty $\regd(\omega \bw)$ goes to infinity with $\omega$.

For any $\bw\in \R^2$ with both components nonzero, there is a
distribution $D$ over $\R^2$ with bounded support such that
the regularization penalty $\regd(\omega \bw)$ goes to infinity with $\omega$.
\end{corollary}

Together Theorems~\ref{t:bounded.regularization} and~\ref{t:two.unbounded} demonstrate that $\regd(\bw)$ is \emph{not} convex
(see also Figure~\ref{f:nonmonotone}).
In fact, $\regd(\bw)$ cannot be approximated to within any factor by a convex function, even if a dependence on $n$ and $p$ is allowed.
For example, Theorem~\ref{t:bounded.regularization}
shows that, for all $D$ with bounded support, both  $\regd(0, \omega)$
and $\regd( \omega, 0 )$ remain bounded 
as $\omega$ goes to infinity, whereas
Theorem~\ref{t:two.unbounded}
shows that there is such a $D$ such that
 $\regd(\omega/2, \omega/2)$ is unbounded as $\omega$ goes to infinity.

Theorem~\ref{t:two.unbounded} relies on the $w_i x_i$ products having different signs.
The following shows that $\regd(\bw)$ does remain bounded when multiple components of $\bw$ go to infinity if the corresponding features are compatible
in the sense that the signs of $w_i x_i$ are always in alignment.

 \begin{theorem}
 \label{t:bounded.mult.reg}
Let $\bw$ be a weight vector and $D$ be a discrete distribution such that    $w_i x_i \geq 0$
for each index $i$ and all $\bx$ in the support of $D$.
The limit of $\regd(\omega \bw)$ as $\omega$ goes to infinity is bounded by  $\ln(2) (1-p) \P_{\bx \sim D}(\bw \cdot \bx \neq 0)$.
\end{theorem}

 The proof of Theorem~\ref{t:bounded.mult.reg}  (which is Appendix~\ref{a:bounded.mult.reg}) easily generalizes to alternative conditions where $\omega \rightarrow - \infty$ and/or
$w_i x_i \leq 0$ for each $i \leq k$ and $\bx$ in the support of $D$.

Taken together Theorems~\ref{t:bounded.mult.reg} and~\ref{t:two.unbounded} give an almost complete characterization of
when multiple weights can go to infinity while maintaining a finite dropout regularization penalty.

\subsection*{Discussion}
The bounds in the  preceding theorems and propositions suggest several properties of the dropout regularizer.
First, the $1-p$  factors indicate that the strength of regularization grows linearly with dropout probability $q=1-p$.
Second, the $\P_{\bx \sim D}(x_i \neq 0)$ factors in several of the bounds suggest that weights for rare features are encouraged by being penalized
less strongly than weights for frequent features.
This preference for rare features is sometimes seen in algorithms like the Second-Order Perceptron \citep{CCG02} and AdaGrad \citep{DHS11}.  \citet{WWL13} discussed the relationship between dropout and these algorithms, based on approximation (8).  Empirical results indicate that dropout performs well in domains like document classification where rare features can have high discriminative value \citep{WM13}.  The theorems of this section suggest that the exact dropout regularizer minimally penalizes the use of rare features.
Finally, Theorem~\ref{t:two.unbounded} suggests that dropout limits co-adaptation by strongly penalizing large weights if
the $w_i x_i$ products often have different signs.
On the other hand, if the $w_i x_i$ products usually have the same sign, then
Proposition~\ref{p:decrease.from.0} indicates that dropout encourages increasing the smaller weights to help share the prediction responsibility.
This intuition is reinforced by Figure~\ref{f:nonmonotone}, where the dropout penalty for two large weights is much less then a single large weight
when the features are highly correlated.

\section{A definition of separation}
\label{s:separation.def}

Now we turn to illustrating the inductive bias of dropout by
contrasting it with $L_2$ regularization.  For this, we will
use a definition of separation between pairs of regularizers.

Each regularizer has a regularization parameter that governs how
strongly it regularizes.  If we want to describe qualitatively what
is preferred by one regularizer over another, we need to control for
the amount of regularization.

Let $\er_P(\bw) = \Pr_{(\bx, y) \sim P}(\sign(\bw \cdot \bx) \neq y)$, and
recall that $\bw^{*}$ and $\bv$ are the minimizers of the dropout and $L_2$-regularized  criteria respectively.

Say that sources $P$ and $Q$ $C$-separate $L_2$ and dropout if
there exist $q$ and $\lambda$ such that both
%
%
$\frac{\er_{P} (\bw^*(P, q))}{\er_{P} (\bv(P, \lambda))} > C$
and $\frac{\er_{Q} (\bv(Q, \lambda))}{\er_{Q} (\bw^*(Q, q))} > C$.
Say that indexed families ${\cal P} = \{ P_{\alpha} \}$
and ${\cal Q}= \{ Q_{\alpha} \}$
{\em strongly separate} $L_2$ and dropout if
pairs of distributions in the family $C$-separate them for
arbitrarily large $C$.  We provide strong separations, using both $n=2$ and larger $n$.


\section{A source preferred by $L_2$}
\label{s:2d.L2.wins}

\newcommand{\PntwoLtwo}{P_{\ref{s:2d.L2.wins}}}

\begin{figure}
  \centerline{
\begin{tabular}{c c}
\includegraphics[width=\figurewidth]{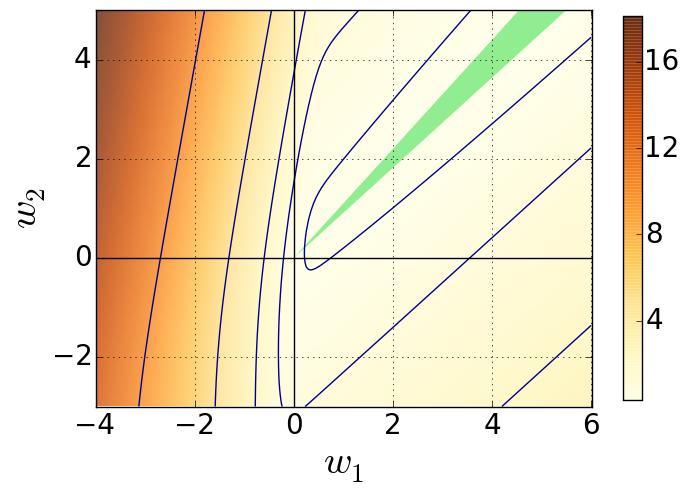} &
\includegraphics[width=\figurewidth]{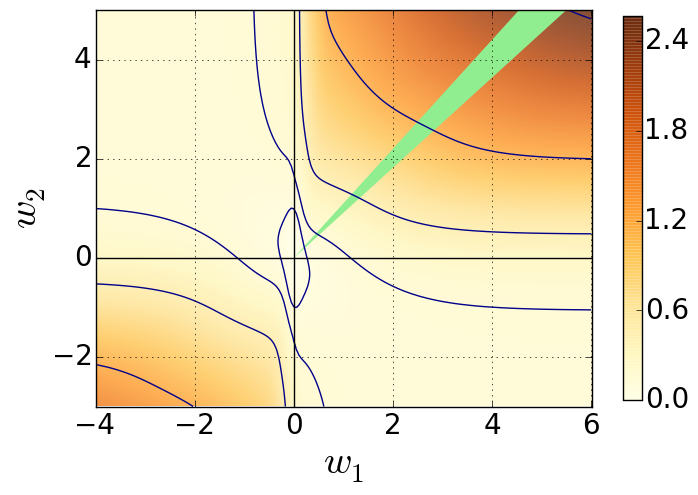} \\
\includegraphics[width=\figurewidth]{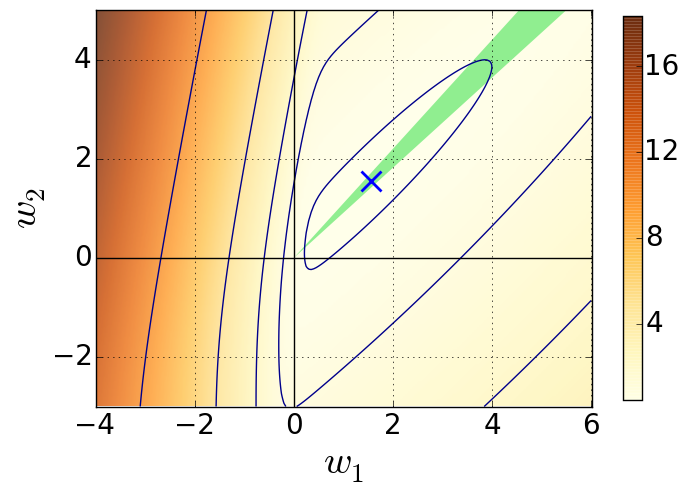} &
\includegraphics[width=\figurewidth]{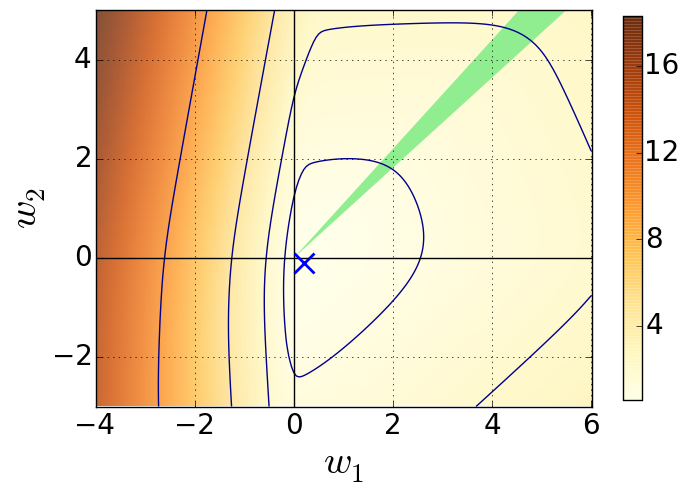}
\end{tabular}
}
\caption{Using data favoring $L_2$ in
  (\ref{e:L2WinsDist}).
The expected loss is plotted in the upper-left, the dropout regularizer in the upper-right,
the  $L_2$ regularized criterion as in~(\ref{e:L2.optimizer}) in the lower-left and the
dropout criterion as in~(\ref{e:dropout.optimizer}) in the lower-right, all as functions of the weight vector.
The Bayes-optimal weight vectors are in the green region, and ``$\times$''  marks show the optimizers of the criteria.}
\label{f:L2WinsPlots}
\end{figure}

Consider the joint distribution $\PntwoLtwo$ defined as follows:
\begin{equation}
\label{e:L2WinsDist}
\begin{array}{cccc}
x_1 & x_2 & y & \Pr(\bx,y) \\ \hline
10 & -1 & 1 & 1/3 \\
1.1 & -1 &  1 & 1/3 \\
-1 & 1.1 & 1 & 1/3 \\
\end{array}
\end{equation}
This distribution has weight vectors that classify examples perfectly (the green shaded region in Figure~\ref{f:L2WinsPlots}).
For this distribution, optimizing an $L_2$-regularized criterion leads to a perfect hypothesis, while
the weight vectors optimizing the dropout criterion make prediction errors on one-third of the distribution.

The intuition behind this behavior for the distribution described in
(\ref{e:L2WinsDist}) is that weight vectors
that are positive multiples of $(1,1)$ classify all of the data correctly.
However, with dropout regularization the $(10,-1)$ and $(1.1, -1)$ data points encourage the second weight to be negative when the first component is dropped out.
This negative push on the second weight is strong enough to prevent the minimizer of the dropout-regularized criterion from correctly classifying the
$(-1, 1.1)$ data point.
Figure~\ref{f:L2WinsPlots} illustrates the loss,  dropout regularization, and dropout and $L_2$ criterion for this data source.


We first show that distribution $\PntwoLtwo$
of (\ref{e:L2WinsDist})
is compatible with mild enough
$L_2$ regularization.
Recall that $\bv(\PntwoLtwo,\lambda)$ is weight vector found by minimizing the $L_2$ regularized criterion~(\ref{e:L2.optimizer}).

\begin{theorem}
\label{t:elltwo.succeeds}
If $0 <  \lambda \leq 1/50$, then
$
\er_{\PntwoLtwo} (\bv(\PntwoLtwo, \lambda)) = 0
$
for the distribution $\PntwoLtwo$ defined in~(\ref{e:L2WinsDist}).
\end{theorem}

In contrast, the $\bw^*(\PntwoLtwo,q)$ minimizing the dropout criterion~(\ref{e:dropout.optimizer})
has error rate at least $1/3$.


\begin{theorem}
\label{t:dropout.fails}
If $q \geq 1/3$ then
$
\er_{\PntwoLtwo} (\bw^*(\PntwoLtwo, q)) \geq 1/3
$
for the distribution $\PntwoLtwo$ defined in~(\ref{e:L2WinsDist}).
\end{theorem}

The proofs of Theorem~\ref{t:elltwo.succeeds} and \ref{t:dropout.fails} are in
Appendices~\ref{a:elltwo.succeeds} and \ref{a:dropout.fails}.

\section{A source preferred by dropout}
\label{s:2d.drop.wins}
%
%
%

\begin{figure}
  \centerline{
\begin{tabular}{c c}
\includegraphics[width=\figurewidth]{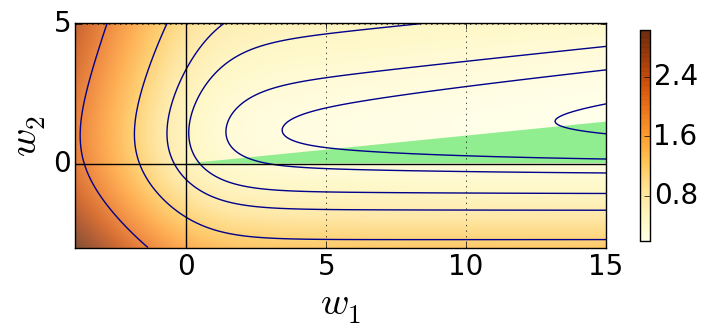} &
\includegraphics[width=\figurewidth]{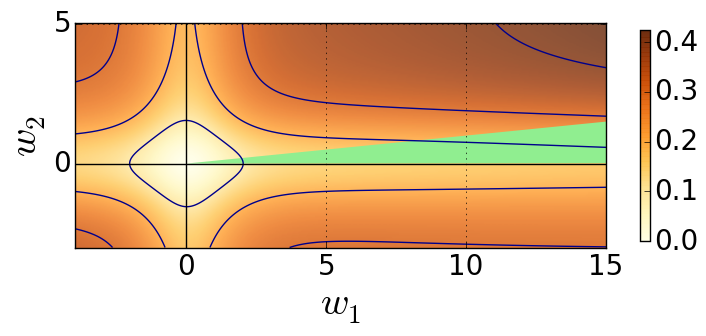} \\
\includegraphics[width=\figurewidth]{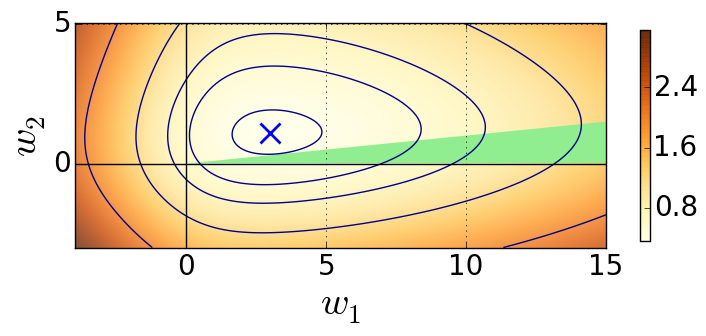} &
\includegraphics[width=\figurewidth]{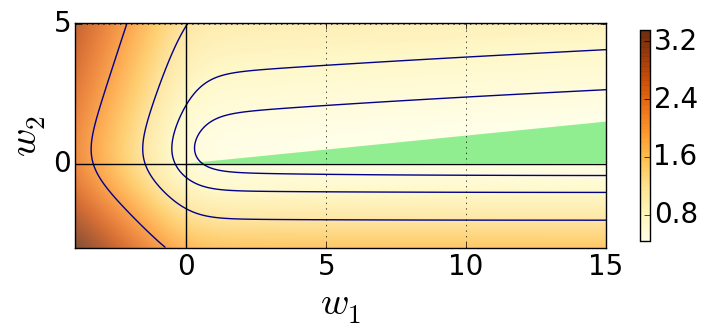} \\
\multicolumn{2}{c}{\includegraphics[width=1.2\figurewidth]{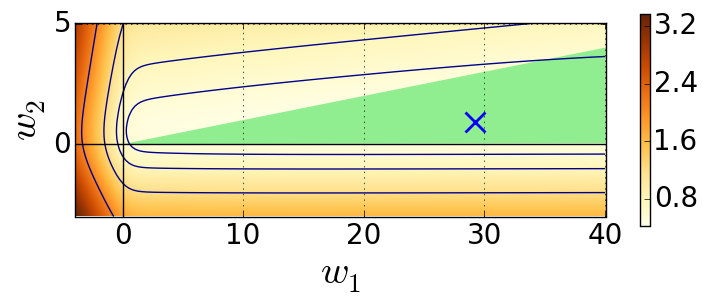}}
\end{tabular}
}

\caption{For the
source from  (\ref{e:drop.wins.dist}) favoring the dropout,
the expected loss is plotted in the upper-left, the dropout regularizer in the upper-right,
the expected loss plus $L_2$ regularization as in~(\ref{e:L2.optimizer}) in the lower-left and the
dropout criterion as in~(\ref{e:dropout.optimizer}) in the lower-right, all as functions of the weight vector.
The Bayes-optimal weight vectors are in the green region, and ``$\times$''  marks show the optimizers of the criteria.
Note that the minimizer of the dropout criterion lies outside the middle-right plot
and is shown on the bottom plot (which has a different range and scale than the others.)}
\label{f:drop.wins.plots}
\end{figure}

\newcommand{\Pntwodrop}{P_{\ref{s:2d.drop.wins}}}
In this section, consider the joint distribution $\Pntwodrop$ defined by
\begin{equation}
  \label{e:drop.wins.dist}
\begin{array}{cccc}
x_1 & x_2 & y & \Pr(\bx,y) \\ \hline
1 & 0 & 1 & 3/7 \\
 -1/1000 & 1 & 1 & 3/7 \\
1/10 & -1 & 1 & 1/7
\end{array}
\end{equation}
The intuition behind this distribution is that the $(1,0)$ data point encourages a large weight on the first feature.
This means that the negative pressure on the second weight due to the $(1/10, -1)$ data point is much smaller (especially given its lower probability)
than the positive pressure on the second weight due to the $(-1/1000, 1)$ example.
The $L_2$ regularized criterion emphasizes short vectors, and prevents the first weight from growing large enough (relative to the second weight)
to correctly classify the $(1/10, -1)$ data point.
On the other hand, the first feature is nearly perfect; it only has the wrong sign on the second example
where it is $-\epsilon = -1/1000$.
This means that, in light of Theorem~\ref{t:bounded.regularization} and Proposition~\ref{p:bounded.reg.arb.w},
dropout will be much more willing to use a large weight for $x_1$, giving it an advantage for this
source over $L_2$.
The plots in Figure~\ref{f:drop.wins.plots} illustrate this intuition.

%

\begin{theorem}
\label{t:elltwo.fails}
If $1/100 \leq \lambda \leq 1$, then
$
\er_{\Pntwodrop} (\bv(\Pntwodrop, \lambda)) \geq 1/7
$
for the distribution $\Pntwodrop$ defined in~(\ref{e:drop.wins.dist}).
\end{theorem}

In contrast, the minimizer of the dropout criterion is able to generalize perfectly.

\begin{theorem}
\label{t:dropout.succeeds}
If $q \leq 1/2$, then
$
\er_{\Pntwodrop} (\bw^*(\Pntwodrop, q)) = 0.
$
for the distribution $\Pntwodrop$ defined in~(\ref{e:drop.wins.dist}).
\end{theorem}

Theorems~\ref{t:elltwo.fails} and \ref{t:dropout.succeeds} are proved in
Appendices~\ref{a:elltwo.fails} and \ref{a:dropout.succeeds}.

\medskip

The results in this and the previous section show that the distributions defined in~(\ref{e:L2WinsDist}) and~(\ref{e:drop.wins.dist}) strongly separate
dropout and $L_2$ regularization.
Theorem~\ref{t:dropout.succeeds} shows that for distribution $P$
analyzed in this section
$\er_P(\bw^*(P, q)) = 0$ for all $q\leq 1/2$ while
Theorem~\ref{t:elltwo.fails} shows that for the same distribution $\er_P(\bv(P,\lambda) \geq 1/7$ whenever $\lambda \geq 1/100$.
In contrast, when $Q$ is the distribution defined
in the previous section,
Theorem~\ref{t:elltwo.succeeds} shows
$\er_Q(\bv(Q,\lambda)) = 0$ whenever $\lambda \leq 1/50$.
For this same distribution $Q$, Theorem~\ref{t:dropout.fails} shows that $\er_Q(\bw^*(Q,q)) \geq 1/3$ whenever $q \geq 1/3$.

\section{$L_1$ regularization}
\label{s:L1}

In this section, we show that the same $\PntwoLtwo$ and $\Pntwodrop$ distributions that separate dropout from $L_2$ regularization 
also separate dropout from $L_1$ regularization:  the
algorithm the minimizes
\begin{equation}
\label{e:ellone.objective}
\E_{(\bx,y) \sim P}
     ( \ell(y \bw \cdot \bx))
    + \lambda || \bw ||_1.
\end{equation}
 
\begin{figure}
  \centerline{
    \raisebox{-0.5 \height}{\includegraphics[width=\figurewidth]{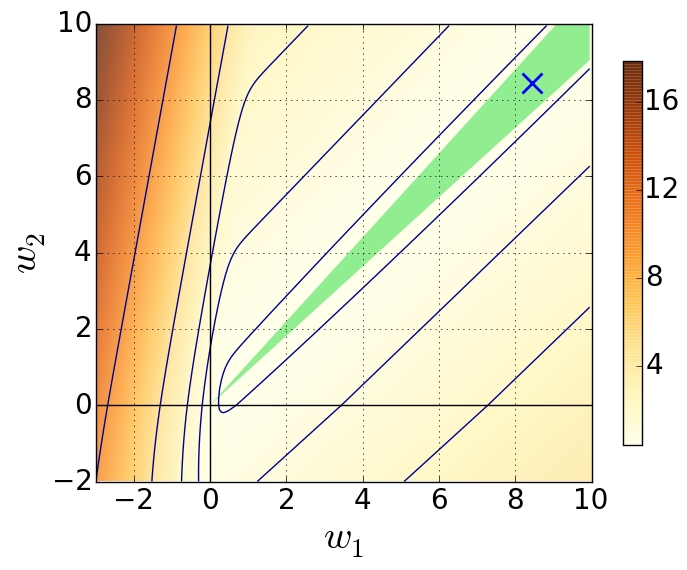}  }
    \qquad  
    \raisebox{-0.5 \height} {\includegraphics[width=\figurewidth]{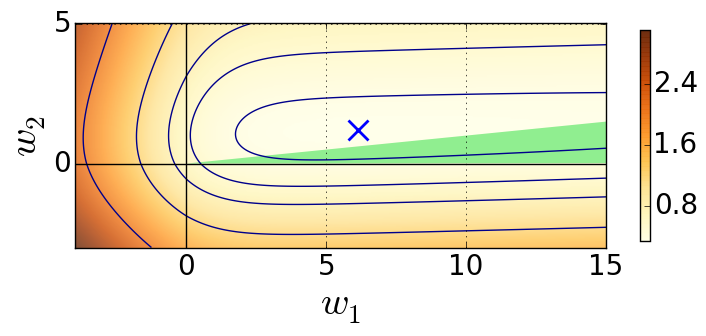} }
    }
  \caption{A plot of the $L_1$ criterion with $\lambda = 0.01$ for distributions $\PntwoLtwo$ defined in Section~\ref{s:2d.L2.wins} (left) 
  	and $\Pntwodrop$ defined in Section~\ref{s:2d.drop.wins} (right).  As before, the Bayes optimal classifiers are denoted by the
  region shaded in green and the minimizer of the criterion is denoted with an x.}
  \label{f:L1.wins}
\end{figure}

As in Sections~\ref{s:2d.L2.wins} and \ref{s:2d.drop.wins}, we set $\lambda = 1/100$.  
Figure~\ref{f:L1.wins} plots the $L_1$ criterion (\ref{e:ellone.objective})
for the distributions $\PntwoLtwo$ defined in~(\ref{e:L2WinsDist}) 
 and $\Pntwodrop$ defined in~(\ref{e:drop.wins.dist}).
Like $L_2$ regularization, $L_1$ regularization produces a Bayes-opitmal classifier on $\PntwoLtwo$, but not on $\Pntwodrop$.
Therefore the same argument shows that these distributions also strongly separate dropout and $L_1$ regularization.

%


\section{A high-dimensional source preferred by $L_2$}
\label{s:L2.manyfeatures}

In this section we exhibit a source where $L_2$ regularization leads to a perfect predictor while dropout regularization creates a predictor with a constant error rate.

\newcommand{\PLtwo}{P_{\ref{s:L2.manyfeatures}}}

Consider the source  $\PLtwo$ defined as follows.
The number $n$ of features is even.
All examples are labeled $1$.
A random example is drawn as follows: the first feature takes the value $1$ with probability $9/10$ and $-1$ otherwise, and
a subset of exactly $n/2$ of the remaining $n-1$ features (chosen uniformly at random) takes the
value $1$, and the remaining $n/2-1$ of those first $n-1$ features
take the value $-1$.

A majority vote over the last $n-1$ features achieves perfect prediction accuracy.
This is despite the first feature (which does not  participate in the vote) being
more strongly correlated with the label than any of the voters in the optimal
ensemble.
Dropout, with its bias for single good features and discrimination against multiple disagreeing features, puts too much weight on this first feature.
In contrast, $L_2$ regularization leads to the Bayes optimal classifier by placing less weight on the first feature than on any of the others.


\begin{theorem}
\label{t:L2.succeeds.large.n}
If $\lambda\leq \frac{1}{30n}$ then  the weight vector $v(\PLtwo, \lambda)$ optimizing the $L_2$ criterion  has perfect prediction accuracy:
$\er_{\PLtwo} (v(\PLtwo, \lambda)) = 0$.
\end{theorem}

When $n>125$, dropout with $q=1/2$ fails to find the Bayes optimal hypothesis.
In particular, we have the following theorem.

\begin{theorem}
\label{t:dropout.fails.large.n}
If the dropout probability $q=1/2$ and the number of features is an even $n > 125$
then the weight vector $\bw^*(\PLtwo, q)$ optimizing the dropout criterion
has prediction error rate $\er_{\PLtwo} (\bw^*(\PLtwo, q)) \geq 1/10$.
\end{theorem}

We conjecture that dropout fails on $\PLtwo$ for all $n \geq 4$.  As evidence, we analyze the $n=4$ case.

\begin{theorem}
\label{t:dropout.fails.n.eq.4}
If dropout probability $q=1/2$ and the number of features is $n =4$
then the minimizer of the dropout criteria $\bw^*(\PLtwo, q)$ has
has prediction error rate $\er_{\PLtwo} (\bw^*(\PLtwo, q)) \geq 1/10$.
\end{theorem}

Theorems~\ref{t:L2.succeeds.large.n}, \ref{t:dropout.fails.large.n} and \ref{t:dropout.fails.n.eq.4} are proved in Appendices~\ref{a:L2.succeeds.large.n}, \ref{a:dropout.fails.large.n} and \ref{a:dropout.fails.n.eq.4}.


\section{A high-dimensional source preferred by dropout}
\label{s:dropout.wins.large.n}

\newcommand{\Pdrop}{P_{\ref{s:dropout.wins.large.n}}}

Define the source $\Pdrop$, which depends on (small) positive real parameters
$\eta$, $\alpha$, and $\beta$, as follows.
A random label $y$ is generated first, with both of
$+1$ and $-1$ equally likely.
The features $x_1,...,x_n$ are
conditionally independent given $y$.
The first feature tends to be accurate but small: $x_1 = \alpha y$ with probability $1 - \eta$, and is $-\alpha y$
 with probability $\eta$.
 The remaining features are larger but less accurate:
for $2\leq i \leq n$, feature $x_i$ is $y$ with probability $1/2 + \beta$,
    and $-y$ otherwise.

When $\eta$ is small enough relative to $\beta$, the Bayes' optimal prediction is to predict with the first feature.
When $\alpha$ is small, this requires concentrating the weight  on $w_1$ to outvote the other features.
Dropout is capable of making this one weight large while $L_2$ regularization  is not.


\begin{theorem}
\label{t:dropout.succeeds.n>1}
If $q = 1/2$, $n \geq 100$, $\alpha > 0$,
$\beta = 1/(10 \sqrt{n-1})$, and
$\eta \leq \frac{1}{2 + \exp(54 \sqrt{n})}$,
then
$
\er_{\Pdrop} (\bw^*(\Pdrop, q)) = \eta.
$
\end{theorem}

%

\begin{theorem}
\label{t:elltwo.fails.n>1}
If $\beta = 1/(10 \sqrt{n-1})$,
$\lambda = \frac{1}{30n}$,
$\alpha < \beta \lambda$,
and $n$ is a large enough even number, then for any $\eta \in [0,1]$,
$
\er_{\Pdrop} (\bv(\Pdrop, \lambda)) \geq 3/10.
$
\end{theorem}

Theorems~\ref{t:dropout.succeeds.n>1} and \ref{t:elltwo.fails.n>1} are
proved in Appendices~\ref{a:dropout.succeeds.n>1} and
\ref{a:elltwo.fails.n>1}.


Let $\tilde n$ be a large enough even number in the sense of  Theorem~\ref{t:elltwo.fails.n>1}.
Let $P_{\eta}$ be the distribution defined at the start of Section~\ref{s:dropout.wins.large.n} with number of features $n = \tilde n$,
$\beta = 1/(10 \sqrt{n-1})$, $\alpha = 1/(300 n \sqrt{n})$,  and $0 < \eta < 1/ (2 + \exp(54 \sqrt{n}))$ is a free parameter.
Theorem~\ref{t:dropout.succeeds.n>1} shows that $\er_{P_\eta}(\bw^*(P_\eta,q)) = \eta$ when dropout probability $q=1/2$.
For this same distribution,
Theorem~\ref{t:elltwo.fails.n>1} shows $\er_{P_\eta} (\bv(P_\eta,\lambda)) \geq 3/10$ when $\lambda = 1/30 n$.
Therefore \[
\frac{\er_{P_\eta}(\bw^*(P_\eta,1/2)) }{\er_{P_\eta} (\bv(P, 1/30 \tilde n))}
\]
goes to 0 as $\eta \rightarrow 0$.

The distribution defined at the start of Section~\ref{s:L2.manyfeatures},
which we call $Q$ here, provides contrasting behavior when $n=\tilde n$.
Theorem~\ref{t:dropout.fails.large.n} shows that the error $\er_Q (\bw^*(Q,1/2) ) \geq 1/10$
while Theorem~\ref{t:L2.succeeds.large.n} shows that $\er_Q(v(Q, 1/30 \tilde n) = 0$.
Therefore the $P_\eta$ and $Q$ distributions
strongly separate dropout and $L_2$ regularization for parameters $q= 1/2$ and $\lambda = 1/30 n$.

\section{Conclusions}
\label{s:conclusions}

We have  built  on the  interpretation of dropout as a regularizer in \citet{WWL13}
to prove  several interesting properties of the dropout regularizer.
This interpretation decomposes the dropout criterion minimized by training into a loss term plus a
regularization penalty that depends on the feature vectors in the training set (but not the labels).
We started with a characterization of when the dropout criterion has a unique minimum, and then
turn to properties of the dropout regularization penalty.
We verified that the dropout regularization penalty has some desirable properties of a regularizer:
it is 0 at the zero vector, and the contribution of each feature vector in the training set is non-negative.

On the other hand, the dropout regularization penalty does not behave like standard regularizers.
In particular, we have shown:
\begin{enumerate}
\item Although the  dropout ``loss plus regularization penalty'' criterion is convex in the weights $\bw$,
the regularization penalty imposed by dropout training is \emph{not} convex.
\item Starting from an arbitrary weight vector, any single weight can go to infinity while the dropout regularization penalty remains bounded.
\item
\label{i:stay.bounded} In some cases, multiple weights can simultaneously go to infinity while the regularization penalty remains bounded.
\item The regularization penalty can \emph{decrease} as weights increase from 0 when the features are correlated. 
\end{enumerate}
These are in stark contrast to standard norm-based regularizers that always diverge as any weight goes to infinity, and are non-decreasing in each individual weight.

In most cases the dropout regularization penalty \emph{does} diverge as multiple weights go to infinity.
We  characterize when sending two weights to infinity
causes the dropout regularization penalty to diverge, and when it will remain finite.
In particular, dropout is willing to put a large weights on multiple features if the $w_i x_i$ products tend to have the same sign.

The form of our analytical bounds suggest that 
 the strength of the regularizer grows linearly with the dropout probability $q$, 
 and provide additional support for the claim \citep{WWL13} that dropout favors rare features.

We found it important to check our intuition by working through small examples.
To make this more rigorous we needed a definition of when a source favored dropout regularization over a more standard regularizer like $L_2$.
Such a definition needs to deal with the strength of regularization, a difficulty complicated by the fact that dropout regularization is parameterized by the dropout probability $q\in [0,1]$ while
$L_2$ regularization is parameterized by $\lambda \in [0, \infty]$.
Our solution is to consider pairs of sources $P$ and $Q$.
We then say the pair \emph{separates} the
dropout and $L_2$ if dropout with a particular parameter $q$ performs better then $L_2$ with a particular parameter $\lambda$ on source $P$, while
$L_2$ (with the same $\lambda$) performs better than dropout (with the same $q$) on source $Q$.
Our definition uses generalization error as the most natural interpretation of ``performs better''.

Sections~\ref{s:2d.L2.wins} through~\ref{s:dropout.wins.large.n} are devoted to proving that dropout and $L_2$ are strongly separated by certain pairs of distributions.  
Section~\ref{s:L1} shows that dropout and $L_1$ regularization are also strongly separated.
Proving strong separation is non-trivial even after one finds the right distributions.
This is due to several factors:  the minimizers of the criteria do not have closed forms, we wish to prove separation for
ranges of the regularization values, and the binomial distributions induced by dropout are not amenable to exact analysis.
Despite these difficulties, the separation results reinforce the intuition that dropout is more willing to use a large weight in order to better fit the training data
than $L_2$ regularization.
However, if two features often have both the same and different signs (as in Theorem~\ref{t:two.unbounded}) then dropout is less willing to put even  moderate weight on both features.

As a side benefit of these analyses, the plots in Figure~\ref{f:L2WinsPlots} and Figure~\ref{f:drop.wins.plots} provide a dramatic illustration of the
dropout regularizer's non-convexity and its preference for making only a single weight large.
This is consistent with the insight provided by Theorems~\ref{t:two.unbounded} and~\ref{t:bounded.mult.reg}.

Our analysis is for the logistic regression case corresponding to a single output node.
It would be very interesting to have similar analysis for multi-layer neural networks.
However, dealing with non-convex loss of such networks will be a major challenge.
Another open problem suggested by this work is how the definition of separation can be used to gain insight about other regularizers and settings.




\bibliography{general}

\begin{thebibliography}{24}
\providecommand{\natexlab}[1]{#1}
\providecommand{\url}[1]{\texttt{#1}}
\expandafter\ifx\csname urlstyle\endcsname\relax
  \providecommand{\doi}[1]{doi: #1}\else
  \providecommand{\doi}{doi: \begingroup \urlstyle{rm}\Url}\fi

\bibitem[Abernethy et~al.(2014)Abernethy, Lee, Sinha, and Tewari]{AbeEtAl14}
J.~Abernethy, C.~Lee, A.~Sinha, and A.~Tewari.
\newblock Online linear optimization via smoothing.
\newblock \emph{COLT}, pages 807--823, 2014.

\bibitem[Bachman et~al.(2014)Bachman, Alsharif, and Precup]{BAP14}
P.~Bachman, O.~Alsharif, and D.~Precup.
\newblock Learning with pseudo-ensembles.
\newblock \emph{NIPS}, 2014.

\bibitem[Baldi and Sadowski(2013)]{BS13}
P.~Baldi and P.~J. Sadowski.
\newblock Understanding dropout.
\newblock In \emph{Advances in Neural Information Processing Systems}, pages
  2814--2822, 2013.

\bibitem[Bartlett et~al.(2006)Bartlett, Jordan, and McAuliffe]{BJM06}
P.~L. Bartlett, M.~I. Jordan, and J.~D. McAuliffe.
\newblock Convexity, classification, and risk bounds.
\newblock \emph{Journal of the American Statistical Association}, 101\penalty0
  (473):\penalty0 138--156, 2006.

\bibitem[Breiman(2004)]{Bre04}
L.~Breiman.
\newblock Some infinity theory for predictor ensembles.
\newblock \emph{Annals of Statistics}, 32\penalty0 (1):\penalty0 1--11, 2004.

\bibitem[Cesa-Bianchi et~al.(2002)Cesa-Bianchi, Conconi, and Gentile]{CCG02}
N.~Cesa-Bianchi, A.~Conconi, and C.~Gentile.
\newblock A second-order perceptron algorithm.
\newblock \emph{COLT}, 2002.

\bibitem[Dahl(2012)]{Dah12}
G.~Dahl.
\newblock Deep learning how i did it: Merck 1st place interview, 2012.
\newblock http://blog.kaggle.com.

\bibitem[Dahl et~al.(2013)Dahl, Sainath, and Hinton]{DSH13}
G.~E. Dahl, T.~N. Sainath, and G.~E. Hinton.
\newblock Improving deep neural networks for lvcsr using rectified linear units
  and dropout.
\newblock \emph{ICASSP}, 2013.

\bibitem[DasGupta(2008)]{Das08}
A.~DasGupta.
\newblock \emph{Asymptotic theory of statistics and probability}.
\newblock Springer, 2008.

\bibitem[Duchi et~al.(2011)Duchi, Hazan, and Singer]{DHS11}
J.~Duchi, E.~Hazan, and Y.~Singer.
\newblock Adaptive subgradient methods for online learning and stochastic
  optimization.
\newblock \emph{JMLR}, 12:\penalty0 2121--2159, 2011.

\bibitem[Graham et~al.(1989)Graham, Knuth, and Patashnik]{GKP89}
R.~L. Graham, D.~E. Knuth, and O.~Patashnik.
\newblock \emph{Concrete Mathematics}.
\newblock Addison-Wesley, 1989.

\bibitem[Helmbold and Long(2012)]{HL12}
D.~P. Helmbold and P.~M. Long.
\newblock On the necessity of irrelevant variables.
\newblock \emph{JMLR}, 13:\penalty0 2145--2170, 2012.

\bibitem[Hinton(2012)]{Hin12}
G.~E. Hinton.
\newblock Dropout: a simple and effective way to improve neural networks, 2012.
\newblock videolectures.net.

\bibitem[Hinton et~al.(2012)Hinton, Srivastava, Krizhevsky, Sutskever, and
  Salakhutdinov]{HSKSS12}
G.~E. Hinton, N.~Srivastava, A.~Krizhevsky, I.~Sutskever, and R.~R.
  Salakhutdinov.
\newblock Improving neural networks by preventing co-adaptation of feature
  detectors, 2012.
\newblock Arxiv, arXiv:1207.0580v1.

\bibitem[Kushner and Yin(1997)]{KY97}
H.~J. Kushner and G.~G. Yin.
\newblock \emph{Stochastic approximation algorithms and applications}.
\newblock Springer, 1997.

\bibitem[L.~Deng(2013)]{DenEtAl13}
e.~a. L.~Deng.
\newblock Recent advances in deep learning for speech research at microsoft.
\newblock \emph{ICASSP}, 2013.

\bibitem[Long and Servedio(2010)]{LS10}
P.~M. Long and R.~A. Servedio.
\newblock Random classification noise defeats all convex potential boosters.
\newblock \emph{Machine Learning}, 78\penalty0 (3):\penalty0 287--304, 2010.

\bibitem[Slud(1977)]{Slu77}
E.~Slud.
\newblock Distribution inequalities for the binomial law.
\newblock \emph{Annals of Probability}, 5:\penalty0 404--412, 1977.

\bibitem[{Van Erven} et~al.(2014){Van Erven}, Kotłowski, and Warmuth]{VKW14}
T.~{Van Erven}, W.~Kotłowski, and M.~K. Warmuth.
\newblock Follow the leader with dropout perturbations.
\newblock \emph{Annual ACM Workshop on Computational Learning Theory}, pages
  949--974, 2014.

\bibitem[Wager et~al.(2013)Wager, Wang, and Liang]{WWL13}
S.~Wager, S.~I. Wang, and P.~Liang.
\newblock Dropout training as adaptive regularization.
\newblock \emph{NIPS}, 2013.

\bibitem[Wager et~al.(2014)Wager, Fithian, Wang, and Liang]{WFWL14}
S.~Wager, W.~Fithian, S.~Wang, and P.~S. Liang.
\newblock Altitude training: Strong bounds for single-layer dropout.
\newblock \emph{NIPS}, 2014.

\bibitem[Wan et~al.(2013)Wan, Zeiler, Zhang, Cun, and Fergus]{WanEtAl13}
L.~Wan, M.~Zeiler, S.~Zhang, Y.~L. Cun, and R.~Fergus.
\newblock Regularization of neural networks using dropconnect.
\newblock In \emph{ICML}, pages 1058--1066, 2013.

\bibitem[Wang and Manning(2013)]{WM13}
S.~Wang and C.~Manning.
\newblock Fast dropout training.
\newblock In \emph{ICML}, pages 118--126, 2013.

\bibitem[Zhang(2004)]{Zha04convex}
T.~Zhang.
\newblock Statistical behavior and consistency of classification methods based
  on convex risk minimization.
\newblock \emph{Annals of Statistics}, 32\penalty0 (1):\penalty0 56--85, 2004.

\end{thebibliography}
\bibliographystyle{abbrvnat}

\appendix

\section{Proof of Proposition~\ref{p:decrease.from.0}}
\label{a:decrease.from.0}
\begin{quote}
{\bf Proposition~\ref{p:decrease.from.0}.}  Fix $p = 1/2$, $w_2>0$, and an arbitrary $\bx \in (0,\infty)^2$.  Let $D$ be the distribution concentrated on $\bx$.
Then $\regd(w_1,w_2)$ locally \underline{decreases} as $w_1$ \underline{increases} from $0$.
\end{quote}
First, we show that assuming $\bx = (2,2)$ is without loss of generality.  When $D$ concentrates all
of its probability on a single $\bx$, let us denote $\reg_{D,1/2}$ by $\reg_{\bx,1/2}$.  Since anyplace $w_1$
appears in the expression for $\reg_{\bx,1/2}$, it is multiplied by $x_1$, if we multiply $w_1$ by some constant
$c$ and divide $x_1$ by $c$, we do not change $w_1 x_1$, and therefore do not change $\reg_{\bx,1/2}$.  The same
holds for $w_2$.  Thus
\[
\reg_{\bx,1/2}(\bw) = \reg_{(2,2),1/2}(w_1 x_1/2, w_2 x_2/2).
\]
If we change variables and let $\tilde{w}_1 = w_1 x_1/2$ and $\tilde{w}_2 = w_2 x_2/2$, then since $x_1$ and $x_2$ are both positive,
$\tilde{w}_2$ is positive iff $w_2$ is, and $\reg_{\bx,1/2}(\bw)$ is increasing with $w_1$ iff $\reg_{(2,2),1/2}(\tilde{\bw})$ is increasing
with $\tilde{w}_1$.

We continue assuming $\bx = (2,2)$.  It suffices to show  $\partial \regd(w_1,w_2) / \partial w_1 \vert_{w_1=0} < 0$.  This derivative is
\begin{equation}
\label{e:partial.deriv}
\frac{ 3 e^{w_2} + e^{-3 w_2} - 3e^{-w_2}- e^{3w_2} }
	{2  ( e^{w_2} + e^{-w_2} ) (e^{2 w_2} + e^{-2 w_2} ) } . 
\end{equation}
The sign depends only on the numerator, which is $0$ when $w_2 = 0$.
The derivative of the numerator with respect to $w_2$ is
$3 e^{w_2} - 3 e^{-3 w_2} + 3e^{-w_2} - 3 e^{3w_2}$, which is negative for
$w_2 > 0$, since $e^z + e^{-z}$ is an increasing function in $z$.
Thus the numerator in~(\ref{e:partial.deriv}) is decreasing in $w_2$.
Therefore~(\ref{e:partial.deriv})  is negative when $w_2>0$, and the regularization penalty is (locally) decreasing 
as $w_1$ increases from 0.

(Note: Proposition~\protect\ref{p:decrease.from.0} may be generalized with slight modifications to apply whenever $\bx$ has two nonzero components.  What is needed is that
$x_1 w_1$ and $x_2 w_2$ have the same sign.  For example, if $x_1$ is negative but $x_2 w_2$ is positive, then moving $w_1$ from $0$ in the negative direction decreases
$\regd(\bw)$.)

\section{Proof of Theorem~\protect\ref{t:two.unbounded}}
\label{a:two.unbounded}
\begin{quote}
{\bf Theorem~\protect\ref{t:two.unbounded}.}
Fix an arbitrary distribution $D$ with support in $\R^2$, weight vector $\bw \in \R^2$, and non-dropout probability $p$.
If there is an $\bx$ with positive probability
under $D$ such that $w_1 x_1 $ and $w_2 x_2$ are both non-zero and have different signs,
then the regularization penalty $\regd(\omega \bw)$ goes to infinity as $\omega$ goes to $\pm \infty$.
\end{quote}
Fix an $\bx$ satisfying the conditions of the theorem.
\begin{align}
\regd (\omega \bw) 
&\geq   D(\bx) \E_{\bnu} \left(
	 \ln \left( \frac{\exp(-\frac{\omega  \bw\cdot (\bx + \bnu )}{2}) +\exp(\frac{\omega \bw \cdot (\bx + \bnu)}{2})}%
	 		{\exp(\frac{-\omega \bw \cdot \bx}{2})+\exp(\frac{\omega \bw \cdot \bx}{2} )} \right)
	 \right)  \nonumber \\
&>  D(\bx)  \E_{\bnu} \left(
	 \ln \left( \frac{\exp(\frac{|\omega  \bw\cdot (\bx + \bnu )|}{2})}{ 2\exp(\frac{|\omega \bw \cdot \bx|}{2})} \right)
	 \right) \nonumber \\
&=  D(\bx) \E_{\bnu} \left(  - \ln (2) + \frac{|\omega  \bw\cdot (\bx + \bnu )|}{2} - \frac{|\omega \bw \cdot \bx|}{2}
	 \right). \label{e:two.unbounded.mid}
 \end{align}
 We now examine the expectation over $\bnu$ of the term that depends on $\bnu$.
 We assume that $|w_1 x_1| \geq |w_2 x_2|$ so $|\bw \cdot \bx| = |w_1 x_1| - |w_2 x_2|$; the other case is symmetrical.
  \begin{align*}
\E_{\bnu} ( |\omega  \bw\cdot (\bx + \bnu )| ) 
&= |\omega| \big( p^2 | \bw \cdot \bx / p| + p (1-p) |w_1 x_1 / p |
   + p(1-p) |w_2 x_2  /p | \big) \\
   &= |\omega| \big( p | \bw \cdot \bx | + (1-p) (|w_1 x_1 | - | w_2 x_2| + |w_2 x_2|)
   + (1-p) |w_2 x_2  | \big) \\
&= |\omega| ( | \bw \cdot \bx | + 2 (1-p)  | w_2 x_2|  ).
 \end{align*}
 Plugging this into~(\ref{e:two.unbounded.mid}) gives:
 \[
 \regd (\omega \bw)  > D(\bx) \left(- \ln 2 +  (1-p) |\omega|  | w_2 x_2|  \right)
 \]
 which goes to infinity as $\omega$ goes to $\pm \infty$.

 \section{Proof of Theorem~\protect\ref{t:bounded.mult.reg}}
 \label{a:bounded.mult.reg}
 \begin{quote}
{\bf Theorem~\protect\ref{t:bounded.mult.reg}.} Let $\bw$ be a weight vector and $D$ be a discrete distribution such that    $w_i x_i \geq 0$
for each index $i$ and all $\bx$ in the support of $D$.
The limit of $\regd(\omega \bw)$ as $\omega$ goes to infinity is bounded by  $\ln(2) (1-p) \P_{\bx \sim D}(\bw \cdot \bx \neq 0)$.
 \end{quote}
First note that If $\bw$ and $D$ are such that $\bw \cdot \bx=0$ for all $\bx$ in the support of $D$,
then $\regd(\bw) = \regd (\omega \bw) = 0$.  We now analyze the general case.
\begin{align}
\regd (\omega \bw)  \nonumber \
&=  \E_{\bx,\bnu} \left(
	 \ln \left( \frac{\exp(\frac{\omega \bw \cdot (\bx + \bnu )}{2})+\exp(\frac{-\omega \bw \cdot (\bx + \bnu)}{2})}
	 			{\exp(\frac{\omega \bw \cdot \bx}{2})+\exp(\frac{-\omega \bw \cdot \bx}{2})} \right)
	 \right) \nonumber \\
&=  \E_{\bx,\bnu} \left(
	\ln \left( \frac{\exp(  \frac{ \omega \bw \cdot (\bx + \nu)}{2}) (1 +\exp( - \omega \bw \cdot (\bx + \bnu)))}
		{\exp( \frac{ \omega \bw \cdot  \bx}{2}) (1 +\exp(- \omega \bw \cdot \bx) )} \right)
	\right) \nonumber \\
&=  \E_{\bx,\bnu} \Big(
	\left(  \omega \bw \cdot (\bx + \bnu)/2 \right) + \ln \left( 1 +\exp\left ( - \omega \bw \cdot (\bx + \bnu) \right) \right) \nonumber \\
& \qquad	 - \left(  \omega \bw \cdot \bx / 2 \right) - \ln \left(1 +\exp \left(- \omega \bw \bx \right) \right) \Big) .  \label{e:four.terms}
\end{align}

Of the four terms inside the expectation in Equation~(\ref{e:four.terms}), the first and third cancel since the expectation of $\bnu$ is $\bzero$.
Therefore:
%
\begin{align}
\label{e:twoterm}
\regd(\omega \bw)  =  \E_{\bx} \big( \E_\bnu \big( 
	\ln ( 1 +\exp ( - \omega \bw \cdot (\bx + \bnu) ) )  -  \ln (1 +\exp (- \omega \bw \bx ) )\big) \big) .
\end{align}

\underline{Define $\nez(\bw, \bx)$} to be the number of indices $i$ where $w_i x_i \neq 0$.
We now consider cases based on $\nez(\bw, \bx)$.

Whenever $\nez(\bw, \bx) = 0$ then both $\bw \cdot \bx = 0$ and $\bw \cdot (\bx + \bnu) = 0$.
Therefore the contribution of these $\bx$ to the expectation in (\ref{e:twoterm}) is
$\ln(2) - \ln (2) = 0$.

If $\nez(\bw, \bx) > 0$ then $\bw \cdot \bx > 0$ (since each $w_i x_i \geq 0$), and the second term of~(\ref{e:twoterm}) goes to zero as $\omega$ goes to infinity.
The first term of~(\ref{e:twoterm}) also goes to zero, \emph{unless} all of the $\nez(\bw, \bx)$ components where $w_i x_i > 0$ are dropped out.
If they are all dropped out, then the first term becomes $\ln(2)$.
The probability that all $\nez(\bw, \bx)$ non-zero components are simultaneously dropped out is $(1-p)^{\nez(\bw, \bx)}$.
With this reasoning we get from~(\ref{e:twoterm}) that:
\begin{align}
& \lim_{\omega \rightarrow \infty} \regd(\omega \bw) \nonumber \\
& =  \sum_{k=1}^{n} \P_{\bx \sim D}(\nez(\bw, \bx) = k) \left( \ln(2) (1-p)^{k} \right)  \label{e:nez.bound} \\
& \leq \sum_{k=1}^{n} \P_{\bx \sim D}(\nez(\bw, \bx) = k) \left( \ln(2) (1-p) \right) \nonumber \\
&= \ln(2) (1-p)  \P(\bw \cdot \bx \neq 0) \nonumber
\end{align}
as desired.

(Note that Equation~\ref{e:nez.bound} gives a precise, but more complex expression for the limit.)

\section{Proof of Theorem~\protect\ref{t:elltwo.succeeds}}
\label{a:elltwo.succeeds}
\begin{quote}
{\bf Theorem~\protect\ref{t:elltwo.succeeds}.}
If $0 <  \lambda \leq 1/50$, then
$
\er_{\PntwoLtwo} (\bv(\PntwoLtwo, \lambda)) = 0
$
for the distribution $\PntwoLtwo$ defined in~(\ref{e:L2WinsDist}). 
\end{quote}
To keep the notation clean  let us abbreviate $\PntwoLtwo$ as just $P$ throughout this proof.

By scaling the $L_2$ criterion we can obtain cancellation in the expectation.
Let $\bv$ be weight vector found by minimizing the following $L_2$ regularized criterion $J$:
\begin{equation}
\label{e:L2.wins.L2.crit}
J(\bw) = 3 \left( \E_{(\bx,y) \sim P}
                   ( \ell(y (\bw \cdot \bx))) + (\lambda/2) || \bw ||^2 \right).
\end{equation}
Note the factor of 3 is to simplify the expressions and doesn't affect the minimizing $\bv$.

We will prove Theorem~\ref{t:elltwo.succeeds} with a series of lemmas.

But first, let's take some partial derivatives:
\begin{align}
\label{e:partial.w1}
\frac{\partial J}{\partial w_1}
& = \frac{-10}{1 + \exp(10 w_1 - w_2)}
       + \frac{-1.1}{1 + \exp(1.1 w_1 - w_2)}
       + \frac{1}{1 + \exp(- w_1 + 1.1 w_2)} + 3 \lambda w_1 \\
\label{e:partial.w2}
\frac{\partial J}{\partial w_2}
& = \frac{1}{1 + \exp(10 w_1 - w_2)}
       + \frac{1}{1 + \exp(1.1 w_1 - w_2)}
       + \frac{-1.1}{1 + \exp(- w_1 + 1.1 w_2)} +3  \lambda w_2.
\end{align}

We will repeatedly use the following basic, well-known, lemma.
\begin{lemma}
\label{l:halfspace}
For any convex, differentiable function $\psi$ defined on $\R^n$ with a unique minimum $\bw^*$, for any $\bw \in \R^n$, if
$g(\bw)$ is the gradient of $\psi$ at $\bw$ then $\bw^*$ is contained in the closed halfspace whose separating hyperplane
goes through $\bw$, and whose normal vector is $-g(\bw)$;  i.e., $\bw^* \cdot g(\bw) \leq \bw \cdot g(\bw)$.
Furthermore, if $g(\bw) \neq \bzero$ then $\bw^* \cdot g(\bw) < \bw \cdot g(\bw)$.
\end{lemma}

Now we're ready to start our analysis of $P$.
\begin{lemma}
\label{l:v1.positive}
If $0 \leq \lambda$, the optimizing $v_1$ is positive.
\end{lemma}
{\bf Proof:}
By Lemma~\ref{l:halfspace},
it suffices to show that there is a point $(0,a_2)$ where both $\frac{\partial J}{\partial w_1} \big|_{(0,a_2)} <0$ and
$\frac{\partial J}{\partial w_2} \big|_{(0,a_2)} = 0$.

From Equation~(\ref{e:partial.w1}):
\[
\frac{\partial J}{\partial w_1} \Bigg|_{(0,a_2)} =
 \frac{-11.1}{1 + \exp(- a_2)}
       + \frac{1}{1 + \exp(1.1 a_2)}
\]
and each term is decreasing as $a_2$ increases.
Since it is negative when $a_2 = -2$,  we have $\frac{\partial J}{\partial w_1} \Big|_{(0,a_2)} <0$ for all $a_2 > -2$.  So, to prove the lemma, if
suffices to show that there is a $a_2 \in (-2,\infty)$ such that the other derivative
$\frac{\partial J}{\partial w_2}  \Big|_{(0,a_2)} = 0$.

From equation~(\ref{e:partial.w2}):
\[
\frac{\partial J}{\partial w_2}  \Big|_{(0,a_2)}
 = \frac{2}{1 + \exp(- a_2)}
       + \frac{-1.1}{1 + \exp(1.1 a_2)} + 3  \lambda a_2
 \]
and each term is continuously increasing in $a_2$.
When $a_2 = -2$,
$\frac{\partial J}{\partial w_2}  \Big|_{(0,a_2)}$ is negative.
On the other hand, $\frac{\partial J}{\partial w_2}  \big|_{(0,0)} $
is positive.
Therefore for some $a_2 \in (-2, 0)$ we have
 $\frac{\partial J}{\partial w_2}  \big|_{(0,a_2)}  = 0$
 as desired.
\qed

\begin{lemma}
\label{l:normal.diagonal}
There is a real $a>0$ such that
\[
\frac{ \partial J(\bw) }{ \partial w_1 } \Bigg|_{(a,a)}
  + \frac{ \partial J(\bw) }{ \partial w_2 } \Bigg|_{(a,a)} = 0.
\]
\end{lemma}
{\bf Proof:}  Applying (\ref{e:partial.w1}) and
(\ref{e:partial.w2}), we get
\begin{eqnarray*}
b \defeq
\frac{ \partial J(\bw) }{ \partial w_1 } \Bigg|_{(a,a)}
 + \frac{ \partial J(\bw) }{ \partial w_2 } \Bigg|_{(a,a)}
 = \displaystyle{\frac{-9}{1 + \exp(9 a)}
       + \frac{-0.2}{1 + \exp(a/10)}
       + 6 \lambda a}. \\
\end{eqnarray*}
Since $b$ is negative when $a = 0$ and is a continuous function of $a$,
and $\lim_{a \rightarrow \infty} b > \infty$, the lemma holds.
\qed

\begin{lemma}
\label{l:v1geqv2}
$v_1 \geq v_2$.
\end{lemma}
{\bf Proof:}  Let $a$ be the value from Lemma~\ref{l:normal.diagonal},
and let $\bg = (g_1,g_2)$ be the gradient of $J$ at $(a,a)$.
Lemma~\ref{l:halfspace} implies that $\bv$ lies in the halfspace through $(a,a)$
in the direction of $-\bg$.
Lemma~\ref{l:normal.diagonal}
implies that
\[
g_1 = \frac{ \partial J(\bw) }{ \partial w_1 } \Bigg|_{(a,a)}
  =  - \frac{ \partial J(\bw) }{ \partial w_2 } \Bigg|_{(a,a)} = -g_2.
\]
Examination of the derivatives~(\ref{e:partial.w1}) and~(\ref{e:partial.w2}) at $(a,a)$ shows that the first term of~(\ref{e:partial.w1}) is negative
and the first term of~(\ref{e:partial.w2}) is positive while the last three terms match (although in a different order).
Therefore $g_1 < 0$ and $g_2 = -g_1$ is positive.  Applying Lemma~\ref{l:halfspace} completes the proof.
\qed

Lemma~\ref{l:v1geqv2} implies that $\bv$ correctly classifies
$(10,-1)$ and $(11/10,-1)$.  It remains to show that $\bv$
correctly classifies $(-1,11/10)$, that is, that $v_1$ is
not {\em too} much bigger than $v_2$.

\bigskip

\begin{lemma}
If  $v_2 \geq 0.6$ and $\lambda > 0$ then $v_1 < 11 v_2 / 10$.
\end{lemma}
{\bf Proof:}
Combining
$\frac{\partial J}{\partial w_1} \Big|_{\bv} = 0$ with (\ref{e:partial.w1}),
we get
\[
3 \lambda v_1 = \frac{10}{1 + \exp(10 v_1 - v_2)}
              + \frac{1.1}{1 + \exp(1.1 v_1 - v_2)}
              + \frac{-1}{1 + \exp(- v_1 + 1.1 v_2)}
\]
and, similarly,
\[
3 \lambda v_2 = \frac{-1}{1 + \exp(10 v_1 - v_2)}
       + \frac{-1}{1 + \exp(1.1 v_1 - v_2)}
       + \frac{1.1}{1 + \exp(- v_1 + 1.1 v_2)}.
\]
Thus
\begin{equation}
 \label{e:lambda.vs.rest}
3 \lambda (10 v_1 - 11 v_2)
=  \frac{111}{1 + \exp(10 v_1 - v_2)}
	   + \frac{22}{1 + \exp(1.1 v_1 - v_2)}
              - \frac{22.1}{1 + \exp(- v_1 + 1.1 v_2)}.
\end{equation}
Assume for contraction that
$v_1 \geq 11 v_2 / 10$.
Then  $10v_1 - v_2 \geq 10 v_2$,
$1.1v_1-v_2 \geq 0.21v_2 $,  and
$-v_1 + 1.1v_2 \leq 0$, so
\[
3 \lambda (10 v_1 - 11 v_2)
\leq  \frac{111}{1 + \exp(10 v_2)}
	   + \frac{22}{1 + \exp(0.21  v_2)}
              - 11.05.
\]
However, $10v_1 - 11 v_2 \geq 0$ and (since $v_2 \geq 0.6$) the RHS is negative, giving the desired
contradiction.
\qed

\begin{lemma}
If $0< \lambda \leq 1/50$ then $v_2 \geq 0.6$.
\end{lemma}
{\bf Proof:}
It suffices to show that there is a point $(x, 0.6)$ where the partial w.r.t.\ $w_1$ is 0 and the partial w.r.t $w_2$
is negative.
\[
\frac{\partial J}{\partial w_1} \Big|_{(x, 0.6) }
 = \frac{-10}{1 + \exp(10 x - 0.6)}
       + \frac{-1.1}{1+ \exp(1.1 x - 0.6 )}
       + \frac{1}{1 + \exp(-x + 0.66)} +  3 \lambda x
\]
and is increasing in $x$ and $\lambda$ (assuming $x > 0$) and becomes positive as $x$ goes to infinity.
It is negative when evaluated at $x = 0.6$ and $\lambda = 1/50$, so for all $\lambda \leq 1/50$ there is an $ x > 1$
such that $\partial J / \partial w_+ \big|_{(x, 1)} = 0$.

\[
\frac{\partial J}{\partial w_2} \Big|_{(x, 0.6) }
 = \frac{1}{1 + \exp(10 x - 0.6)}
       + \frac{1}{1 + \exp(1.1 x - 0.6)}
       + \frac{-1.1}{1 + \exp(- x + 0.66)} + 1.8 \lambda \\
\]
and is decreasing in $x$ and increasing in $\lambda$.
It is negative when $x = 1$ and $\lambda=1/50$, so it will remain negative for all $x > 1$ and $0 \leq \lambda \leq 1/50$,
as desired.
\qed

\medskip

So, we have shown that, if $\lambda \leq 1/50$, then all examples are
classified correctly by $\bv$, which proves Theorem~\ref{t:elltwo.succeeds}.

\section{Proof of Theorem~\protect\ref{t:dropout.fails}}
\label{a:dropout.fails}
\begin{quote}
{\bf Theorem~\protect\ref{t:dropout.fails}.}
If $q \geq 1/3$ then
$
\er_{\PntwoLtwo} (\bw^*(\PntwoLtwo, q)) \geq 1/3
$
for the distribution $\PntwoLtwo$ defined in~(\ref{e:L2WinsDist}). 
\end{quote}

Throughout this proof we also abbreviate $\PntwoLtwo$ as just $P$.

For this subsection, let us define the scaled dropout criterion
\begin{equation}
\label{e:L2.wins.drop.crit}
J(\bw) = 3 \; \E_{(\bx,y) \sim P, \br}
                  ( \ell(y (\bw \cdot (\br \odot \bx))))
\end{equation}
where the components of $\br$ are independent
samples from a Bernoulli distribution with parameter $p=1-q>0$.
Again, the factor of 3 is to simplify the expectation and doesn't change the minimizing $\bw$.
Let $\bw^{\circledast}$ be the minimizer of this $J(\bw)$, so that Equation~(\ref{e:r.odot})
implies that the optimizer $\bw^*$ of the dropout criterion is
$p \bw^{\circledast}$.
Note that $\bw^*$ classifies an example correctly if and only if
$\bw^{\circledast}$ does.

Next, note that we may assume without loss of generality that
both components of $\bw^{\circledast}$ are positive, since, if either is negative,
one of $(-1,1.1)$ or $(1.1,-1)$ is misclassified and we are done.

We will prove Theorem~\ref{t:dropout.fails} by proving that, when
$q \geq 1/3$, $\bw^{\circledast}$ misclassifies $(-1,1.1)$, or,
equivalently, that $w^{\circledast}_1 > (11/10) w^{\circledast}_2$.

First, let us evaluate some partial derivatives.  (Note that, if $x_i$
is dropped out, the value of $w_i$ does not matter.)
\begin{align}
\label{e:partial.w1.do}
\frac{\partial J}{\partial w_1}
& = (1-q)^2 \left( \frac{-10}{1 + \exp(10 w_1 - w_2)}
                  + \frac{-1.1}{1 + \exp(1.1 w_1 - w_2)}
                  + \frac{1}{1 + \exp(- w_1 + 1.1 w_2)} \right) \\
\nonumber
& \hspace{0.2in} +
     (1-q) q \left( \frac{-10}{1 + \exp(10 w_1)}
                    + \frac{-1.1}{1 + \exp(1.1 w_1)}
                    + \frac{1}{1 + \exp(- w_1)} \right) \\
\label{e:partial.w2.do}
\frac{\partial J}{\partial w_2}
& = (1-q)^2 \left( \frac{1}{1 + \exp(10 w_1 - w_2)}
              + \frac{1}{1 + \exp(1.1 w_1 - w_2)}
              + \frac{-1.1}{1 + \exp(- w_1 + 1.1 w_2)} \right) \\
\nonumber
& \hspace{0.2in} +
 q (1-q) \left( \frac{1}{1 + \exp(- w_2)}
               + \frac{1}{1 + \exp(- w_2)}
               + \frac{-1.1}{1 + \exp(1.1 w_2)} \right).
\end{align}

The following is the key lemma.  As before, it is useful since,
for any $\bw$, if $g(\bw)$ is nonzero, then $\bw^{\circledast}$ lies in the open halfspace
through $\bw$ whose normal vector is the negative gradient.
\begin{lemma}
\label{l:bad.halfspace}
For all $a > 0$ and $q \geq 1/3$,
\begin{equation}
\label{e:correct.side}
\frac{\partial J}{\partial w_2} \Bigg|_{(a, 10 a/11)} > 0.
\end{equation}
\end{lemma}
{\bf Proof:}  We have
\begin{align*}
\frac{\partial J}{\partial w_2} \Bigg|_{(a, 10 a/11)}
 = & (1-q)^2 \left( \frac{1}{1 + \exp(100 a/11)}
              + \frac{1}{1 + \exp(21 a/110)}
              + \frac{-1.1}{2} \right) \\
\nonumber
& +
 q (1-q) \left( \frac{2}{1 + \exp(- 10a/11)}
               + \frac{-1.1}{1 + \exp(a)} \right).
\end{align*}
Note that this derivative is positive if and only if
\begin{align*}
f(q,a) =
	& \left( \frac{1}{1-q} \right) \; \frac{\partial J}{\partial w_2} \Bigg|_{(a, 10 a/11)}  \\
= 	&  q\left(     \frac{11}{20} +  \frac{2}{1+\exp(-10a/11) } +  \frac{-1}{1 + \exp(21 a/110)}  +  \frac{-1}{1 + \exp(100a/11)}  +  \frac{-11/10}{1+\exp(a)}  \right) \\
	& \hspace{0.2in} + \frac{1}{1+\exp(21a / 110)} + \frac{1}{1+\exp(100a / 11)} + \frac{-11}{20}
\end{align*}
is positive, as $0 < q < 1$.
Note that the terms multiplying $q$ are increasing in $a$ and sum to 0 when $a=0$.
On the other hand, the terms not multiplied by $q$ are decreasing in $a$ and turn negative when $a$ is just over $1/4$.
Thus both parts are positive when $a \leq 1/4$.
Note that $f(q,a)$ can be underestimated by  underestimating $a$ on the $q$-terms and overestimating $a$ on the other terms.

For $1/4 \leq a \leq 2$,
\begin{align*}
f(q,a)  \geq
 & q\left(     \frac{11}{20} +  \frac{2}{1+\exp(-10/44) } +  \frac{-1}{1 + \exp(21 /440)}  +  \frac{-1}{1 + \exp(100/44)}  +  \frac{-11/10}{1+\exp(1/ 4)}  \right) \\
& + \frac{1}{1+\exp(42 / 110)} + \frac{1}{1+\exp(200 / 11)} + \frac{-11}{20} \\
& \geq 0.5 q  - 0.15
\end{align*}
and is positive whenever $q \geq 1/3$.

For $a \geq 2$,
\begin{align*}
f(q,a)
& \geq   q\left(     \frac{11}{20} +  \frac{2}{1+\exp(-20/11) } +  \frac{-1}{1 + \exp(42/110)}  +  \frac{-1}{1 + \exp(200/11)}  +  \frac{-11/10}{1+\exp(2)}  \right)
 + \frac{-11}{20} \\
& \geq  1.7 q - 11/20
\end{align*}
and is also positive whenever $q\geq 1/3$.  \qed

\medskip

{\bf Proof of Theorem~\ref{t:dropout.fails}}:
Let $\bg  = (g_1, g_2)$ be the gradient of $J$ at $(w_1^{\circledast}, 10 w_1^{\circledast} /11)$.
Lemma~\ref{l:bad.halfspace} shows $\bg$ is not ${\bf 0}$, so by convexity
\[
\bw^{\circledast} \cdot \bg  <
(w_1^{\circledast}, 10 w_1^{\circledast} /11) \cdot \bg
\]
which implies
\[
w_2^{\circledast} \; g_2 < (10 w_1^{\circledast}/11) \; g_2.
\]
Since $g_2 > 0$ (Lemma~\ref{l:bad.halfspace}),  this implies
\[
w_2^{\circledast}  < (10 w_1^{\circledast}/11)
\]
and the $(-1, 11/10)$ example is misclassified by $\bw^{\circledast}$, and therefore
by $\bw^*$, completing the proof. \qed

\section{Proof of Theorem~\ref{t:elltwo.fails}}
\label{a:elltwo.fails}
\begin{quote}
{\bf Theorem~\ref{t:elltwo.fails}.}
If $1/100 \leq \lambda \leq 1$, then
$
\er_{\Pntwodrop} (\bv(\Pntwodrop, \lambda)) \geq 1/7
$
for the distribution $\Pntwodrop$ defined in~(\ref{e:drop.wins.dist}).
\end{quote}

To keep the notation clean, in this section let us abbreviate $\Pntwodrop$ simply as $P$.

As the reader might expect, we will prove Theorem~\ref{t:elltwo.fails}
by proving that $\bv$ fails to correctly classify
$(1/10,-1)$, that is, by proving that $v_1 < 10 v_2$.

We may assume that $v_1 > 0$, since, otherwise, $(1,0)$
is misclassified.

To obtain cancellation in the expectation, we work with the scaled $L_2$ criterion
\begin{equation}
\label{e:drop.wins.L2.crit}
J(\bw) = 7 \E_{(\bx,y) \sim P}
                   ( \ell(y (\bw \cdot \bx))) + (7 \lambda/2) || \bw ||^2.
\end{equation}
and let $\bv(P,\lambda)$ be the vector minimizing this $J$, which
we often abbreviate as simply $\bv$, leaving it implicitly a function of $\lambda$.
Note that this scaling of the criteria does not change the minimizing $\bv$.

Taking derivatives,
\begin{align}
\label{e:partial.w1.elltwo.fails}
\frac{\partial J}{\partial w_1}
& = \frac{-3}{1 + \exp(w_1)}
    + \frac{3 \epsilon}{1 + \exp(-\epsilon w_1 + w_2)}
       + \frac{-0.1}{1 + \exp(w_1/10 - w_2)} + 7  \lambda w_1 \\
\label{e:partial.w2.elltwo.fails}
\frac{\partial J}{\partial w_2}
& = \frac{-3}{1 + \exp(-\epsilon w_1 + w_2)}
       + \frac{1}{1 + \exp(w_1/10 - w_2)} + {7 \lambda w_2}.
\end{align}

\begin{lemma}
\label{l:grad1.positive.elltwo.fails}
If  either:  $\lambda \geq 1/100$ and $a \geq 1/3$,
or $\lambda \geq 1/4$ and $a \geq 1/15$ then
\[
\frac{ \partial J(\bw) }{ \partial w_1 } \Big|_{(10a,a)}  > 0.
\]
\end{lemma}
{\bf Proof}:
We have
\[
\frac{ \partial J(\bw) }{ \partial w_1 } \Bigg|_{(10a,a)}
 = \frac{-3}{(1 + \exp(10 a))}
    + \frac{3 \epsilon}{1 + \exp((1 - 10 \epsilon) a)}
    + \frac{-1}{20} + 70 \lambda a
 > \frac{-3}{(1 + \exp(10 a))}
    + \frac{-1}{20} + 70 \lambda  a.
\]
Each term of the RHS is non-decreasing in $a$ and $\lambda$, and the RHS is positive when
either
$\lambda = 1/100$ and $a = 1/3$ or
$\lambda = 1/4$ and $a = 1/15$.
\qed

To apply this, we want to show that $v_2$ is large enough, which we do next.
\begin{lemma}
\label{l:v2.big.elltwo.fails}
If $\lambda \leq 1/4$ then $v_2 \geq 1/3$ and if $\lambda \leq 1$ then $v_2 \geq 1/15$.
\end{lemma}
{\bf Proof:}  Assume to the contrary that $\lambda \leq 1/4$ but $v_2 < 1/3$.
From (\ref{e:partial.w2.elltwo.fails}), and using that $v_1 > 0$, we have
\begin{equation}
\label{e:partial.bound}
\frac{\partial J}{\partial w_2} \Bigg|_{\bv}
 < \frac{-3}{1 + \exp(v_2)}
       + \frac{1}{1+ \exp(-v_2)}  + {7 \lambda v_2},
\end{equation}
a bound that is increasing in $v_2$ and $\lambda$.
Since $\frac{\partial J}{\partial w_2} \Big|_{\bv} = 0$,  the bound must be positive.
However, when $v_2 \leq 1/3$ and $\lambda \leq 1/4$, it is negative, giving the desired contradiction.

Since the bound~(\ref{e:partial.bound}) is also negative at $v_2 = 1/15$ and $\lambda = 1$,
a similar contradiction proves the other half of the lemma.
\qed

{\bf Proof:} (of Theorem~\ref{t:elltwo.fails}):
Lemmas~\ref{l:grad1.positive.elltwo.fails} and
\ref{l:v2.big.elltwo.fails} imply that
$(10 v_2, v_2)$ is not the minimizing $\bv$ (when $\lambda \geq 1/100$), so by convexity,
\begin{align}
J(10v_2, v_1) + \bigl( (v_1,v_2) - (10v_2, v_2) \bigr) \cdot \nabla J(10 v_2, v_2) &< J(v_1,v_2) \\
(v1 - 10 v_2)\  \frac{\partial J}{\partial w_2} \Bigg|_{(10 v_2, v_2)} & < 0.
 \end{align}

 If $ 1/100 \leq \lambda \leq 1/4$ then Lemma~\ref{l:v2.big.elltwo.fails}
 shows that $v_2 \geq 1/3$ and if
 $1/4 \leq \lambda \leq 1$ then it shows that $v_2 \geq 1/15$.
 In either case, Lemma~\ref{l:grad1.positive.elltwo.fails} shows that
 that $ \frac{\partial J}{\partial w_2} \Big|_{(10 v_2, v_2)}> 0$.
Therefore,
\[
v_1  < 10 v_2
\]
and $(0.1, -1)$ is misclassified by $\bv$, completing
the proof. \qed

\section{Proof of Theorem~\ref{t:dropout.succeeds}}
\label{a:dropout.succeeds}
\begin{quote}
{\bf Theorem~\ref{t:dropout.succeeds}.}
If $q \leq 1/2$, then
$
\er_{\Pntwodrop} (\bw^*(\Pntwodrop, q)) = 0.
$
for the distribution $\Pntwodrop$ defined in~(\ref{e:drop.wins.dist}).
\end{quote}

In this proof, let us abbreviate $\Pntwodrop$ with just $P$, and use $\epsilon$ to denote $1/1000$.
  
\begin{table}
\caption{Seven times the dropout distribution.  The three probability sub-columns correspond to the original examples (1,0), (-1/1000, 1), (1/10, -1), and the final column is the over-estimate  used in Lemma~\ref{l:dropout.wins.w2.big}.}
\label{t:dropout.distribution}

\[
\begin{array}{ | c c | c | c c c | c |}
\hline
{x_1 r_1} & {x_2 r_2} & {y} & \multicolumn{3}{c | }{\mbox{seven times probability} } & \mbox{$\bw^{\circledast} \cdot (\br \odot \bx)$ over-estimate} \\ \hline
 0 & 0 & 1 & 3q &+ 3q^2 & +q^2  	& 0	 \\
 1 & 0 & 1 & 3(1-q) & & 			& \infty	\\
 0 & 1 & 1 & & 3q(1-q) & 			& w_2 \\
 -1/1000 & 0 & 1 & & 3 q (1-q) & 	& 0 \\
  -1/1000 & 1 & 1 & & 3 (1-q)^2 & 	& w_2 \\
  0 & -1 & 1 &  & & q(1-q) 			& \infty \\
1/10 & 0 & 1 & & & q(1-q) 			& \infty \\
1/10 & -1 & 1 & & & (1-q)^2		& \infty  \\ \hline
 \end{array}
 \]
 \end{table}

For this section, let us define the scaled dropout criterion
\begin{equation}
\label{e:drop.wins.drop.crit}
J(\bw) = 7 \E_{(\bx,y) \sim P, \br}
                  ( \ell(y (\bw \cdot (\br \odot \bx)))),
\end{equation}
where, as earlier, the components of $\br$ are independent samples
from a Bernoulli distribution with parameter $p=1-q=1/2>0$.  (Note
that, similarly to before, scaling up the objective function by 7 does
not change the minimizer of $J$.)  See
Table~\ref{t:dropout.distribution} for a tabular representation of the
distribution after dropout.  Let $\bw^{\circledast}$ be the minimizer of $J$, so
that $\bw^* = p \bw^{\circledast}$ (see Equation~(\ref{e:r.odot})).

First, let us evaluate some partial derivatives (note that $1-q = (1-q)^2 + q(1-q)$).
\begin{align}
\label{e:partial.w1.do.succ}
\frac{\partial J}{\partial w_1}
& = (1-q)^2 \left( \frac{-3}{1 + \exp(w_1)}
            + \frac{3 \epsilon}{1 + \exp(-\epsilon w_1 + w_2)}
            + \frac{-0.1}{1 + \exp(0.1 w_1 - w_2)} \right) \\
\nonumber
& \hspace{0.2in} +
     (1-q) q \left(  \frac{-3}{1 + \exp(w_1)}
            + \frac{3 \epsilon}{1 + \exp(-\epsilon w_1)}
            + \frac{-0.1}{1 + \exp(0.1 w_1)} \right) \\
\label{e:partial.w2.do.succ}
\frac{\partial J}{\partial w_2}
& = (1-q)^2 \left( \frac{-3 }{1 + \exp(-\epsilon w_1 + w_2)}
                   + \frac{1}{1 + \exp(0.1 w_1 - w_2)} \right) \\
\nonumber
& \hspace{0.2in} +
 q (1-q) \left( \frac{-3 }{1 + \exp(w_2)}
                   + \frac{1}{1 + \exp(- w_2)} \right).
\end{align}

Let's get started by showing that $\bw^{\circledast}$ correctly classifies
$(1,0)$.
\begin{lemma}
\label{l:w1.positive.dos}
$
w_1^{\circledast} > 0.
$
\end{lemma}
{\bf Proof}:  As before,
it suffices to show that there is a point
$(0,a_2)$ where both $\frac{\partial J}{\partial w_1} \big|_{(0,a_2)} <0$ and
$\frac{\partial J}{\partial w_2} \big|_{(0,a_2)} = 0$.

From Equation~(\ref{e:partial.w1.do.succ}):
\[
\frac{\partial J}{\partial w_1} \Big|_{(0,a_2)}
 = (1-q)^2 \left( \frac{-3}{2}
            + \frac{3 \epsilon}{1 + \exp(a_2)}
            + \frac{-0.1}{1 + \exp(- a_2)} \right)
              +  \frac{(1-q) q }{2} \left( -3.1 + 3\epsilon  \right)
\]
which is decreasing in $a_2$, and negative even as $a_2$ approaches
$-\infty$ (recalling $\epsilon = 1/1000$), so $\frac{\partial J}{\partial w_1} \Big|_{(0,a_2)}$
is always negative.

Equation~(\ref{e:partial.w2.do.succ}) implies
\[
\frac{\partial J}{\partial w_2} \Big|_{(0,a_2)}
 = (1-q)^2 \left( \frac{-3 }{1+\exp(a_2)}  + \frac{1}{1+\exp(-a_2)} \right)
 	+ q(1-q) \left(\frac{-3}{1+\exp(a_2)} + \frac{1}{1+\exp(-a_2)} \right) .
 \]
This is negative when $a_2 = 0$, approaches $1-q$ as $a_2$ goes to
infinity, and is continuous, so there is a $a_2$ such that
$\frac{\partial J}{\partial w_2} \Big|_{(0,a_2)} = 0$.  Since
$\frac{\partial J}{\partial w_1} \Big|_{(0,a_2)} < 0$, this proves the
lemma.
\qed

Next, we'll start to work on showing that $\bw^{\circledast}$ correctly classifies
$(-\epsilon,1)$.
\begin{lemma}
\label{l:second_eg.grad.pos}
For all $a > 1/10$,
\[
\frac{\partial J}{\partial w_1} \Bigg|_{(a/\epsilon,a)} > 0.
\]
\end{lemma}
{\bf Proof}:   From (\ref{e:partial.w1.do.succ}), we have
\begin{align*}
\frac{\partial J}{\partial w_1} \Big|_{(a/\epsilon,a)}   =
	&(1-q)^2  \left ( \frac{-3 }{1+ \exp( a/ \epsilon)} + \frac{3 \epsilon }{ 1 + \exp(0)} + \frac{ -0.1}{ 1+\exp(0.1 (a/ \epsilon) - a)} \right)  \\
	&   + q(1-q) \left( \frac{-3}{1+\exp(a/ \epsilon) } + \frac{3 \epsilon }{ 1 + \exp(-a) }  + \frac{-0.1 }{1+\exp(a / 10 \epsilon)}  \right)
\end{align*}
which is positive if $a > 1/10$ as the positive terms (even with the $\epsilon$ factors) dominate the negative ones.
\qed

\begin{lemma}
\label{l:dropout.wins.w2.big}
\[
w^{\circledast}_2 > 1/4.
\]
\end{lemma}
{\bf Proof:}
Assuming $w_1 \geq 0$, the estimates in Table~\ref{t:dropout.distribution} along with the facts that $\ell(z)$ is positive and decreasing show :
\begin{equation}
\label{e:w2.bound.on.J}
J(\bw) \geq  3 (1-q)  \ln(1+\exp(-w_2)) + 6 q \ln (2) + q^2 \ln (2)
\end{equation}
which is decreasing in $w_2$.
If $w^{\circledast}_2 \leq 1/4$,  then
bound~(\ref{e:w2.bound.on.J}) and the fact that $w^{\circledast}_1 > 0$ (Lemma~\ref{l:w1.positive.dos})  imply that
\[
J(\bw^{\circledast}) \geq 0.69 q^2 + 2.4q + 1.7 .
\]

On the other hand,
\[
J(100,2) \leq - 1.5 q^2 + 6 q + 0.42 ,
\]
and the upper bound on $J(100,2)$  is less than the lower bound on $J(\bw^{\circledast})$ when $0 \leq q \leq 1/2$, giving the desired contradiction.
\qed

Now, we're ready to show that $\bw^{\circledast}$ correctly classifies $(-\epsilon,1)$.
\begin{lemma}
\label{l:dropout.second.correct}
$\epsilon w^{\circledast}_1 < w^{\circledast}_2$.
\end{lemma}
{\bf Proof}:  Let $\bg$ be the gradient of $J$ evaluated at $(w^{\circledast}_2 /  \epsilon, w^{\circledast}_2)$.
Combining Lemmas~\ref{l:second_eg.grad.pos} and \ref{l:dropout.wins.w2.big},
$\bg \neq (0,0)$, so
\[
\bw^{\circledast} \cdot \bg < (w_2^{\circledast}/\epsilon, w_2^{\circledast}) \cdot \bg.
\]
This implies
\[
w^{\circledast}_1  \; \frac{\partial J}{\partial w_1} \Big|_{(w^{\circledast}_2 /  \epsilon, w^{\circledast}_2)}  < \frac{w_2^{\circledast}}{\epsilon } \; \frac{\partial J}{\partial w_1} \Big|_{(w^{\circledast}_2 /  \epsilon, w^{\circledast}_2)} .
\]
Since Lemmas~\ref{l:second_eg.grad.pos} and \ref{l:dropout.wins.w2.big} imply that
$g(w_2^{\circledast}/\epsilon, w_2^{\circledast})_1 > 0$, this completes the proof.  \qed

Finally, we are ready to work on showing that $(1/10,-1)$ is correctly
classified by $\bw^{\circledast}$, i.e.\ that $w_1^{\circledast} > 10 w_2^{\circledast}$.
\begin{lemma}
\label{l:gradient.bound.2.dropout.succeeds}
For all $a \in \R$,
\[
\frac{\partial J}{\partial w_1} \Big|_{(10 a,a)} < 0.
\]
\end{lemma}
{\bf Proof:}  Choose $a \in R$. From (\ref{e:partial.w1.do.succ}),
we have
\begin{align*}
\frac{\partial J}{\partial w_1} \Big|_{(10 a,a)}
= \; & 	 q(1-q) \left( \frac{-3}{1+\exp(10 a)}  + \frac{3 \epsilon}{1+\exp(-10 \epsilon a)} + \frac{-1}{10( 1+\exp(a))} \right) \\
   &		+ (1-q)^2 \left( \frac{-3}{1+\exp(10 a)} + \frac{3 \epsilon}{1+\exp(a-10 \epsilon a)} + \frac{-1}{20} \right) \\
\leq \; & (1-q)^2 \left( 6 \epsilon + \frac{-1}{20} \right) < 0
\end{align*}
using $q \leq 1/2$ and $\epsilon = 1/1000$. \qed

\begin{lemma}
\label{l:dropout.correct.example.3}
$w_1^{\circledast} > 10 w_2^{\circledast}$.
\end{lemma}
{\bf Proof:}  Let $\bg$ be the gradient of $J$ evaluated at $\bu = (10 w^{\circledast}_2, w^{\circledast}_2)$.
Lemma~\ref{l:gradient.bound.2.dropout.succeeds}
implies that $\bg \neq (0,0)$, i.e.\ that
$w_1^{\circledast} \neq 10 w^{\circledast}_2$.  Therefore,
\[
\bw^{\circledast} \cdot \bg < \bu \cdot \bg
\]
which, since $u_2 = w^{\circledast}_2$, implies
\[
w_1^{\circledast}  \frac{\partial J}{\partial w_1} \Big|_{\bu} < 10 w^{\circledast}_2  \frac{\partial J}{\partial w_1} \Big|_{\bu} .
\]
Since Lemma~\ref{l:gradient.bound.2.dropout.succeeds} implies that
${\partial J}/ {\partial w_1} \Big|_{\bu}  < 0$,
this in turn implies
\[
w_1^{\circledast} > 10 w^{\circledast}_2,
\]
completing the proof. \qed

Now we have all the pieces to prove that dropout succeeds on
$P$.

\medskip
{\bf Proof} (of Theorem~\ref{t:dropout.succeeds}):
Lemma~\ref{l:w1.positive.dos} implies that $(1,0)$ is classified
correctly by $\bw^{\circledast}$, and  therefore by $\bw^* = p \bw^{\circledast}$.
Lemma~\ref{l:dropout.second.correct} implies
that $(-\epsilon,1)$ is classified correctly.
Lemma~\ref{l:dropout.correct.example.3} implies that $(1/10,-1)$
is classified correctly, completing the proof. \qed

\section{Proof of Theorem~\ref{t:L2.succeeds.large.n}}
\label{a:L2.succeeds.large.n}
\begin{quote}
{\bf Theorem~\ref{t:L2.succeeds.large.n}.}
If $\lambda\leq \frac{1}{30n}$ then  the weight vector $v(\PLtwo, \lambda)$ optimizing the $L_2$ criterion  has perfect prediction accuracy:
$\er_{\PLtwo} (v(\PLtwo, \lambda)) = 0$.
\end{quote}

In this proof, let us abbreviate $\PLtwo$ as just $P$.

By symmetry and convexity, the optimizing $\bv$ is of the form $(v_1, v_2, v_2, \ldots, v_2)$ with
the last $n-1$ components being equal.
Thus for this distribution minimizing the $L_2$ criterion is equivalent to minimizing the simpler criterion  $K(w_1,w_2)$ defined by:

\[
K(w_1,w_2) = \frac{9}{10} \ln \left(1 + \exp(-w_1 - w_2) \right)
		+ \frac{1}{10} \ln \left(1 + \exp(w_1 -w_2) \right)
		+ \frac{\lambda}{2} \left( w_1^2 + (n-1) w_2^2 \right) .
\]

Let $(v_1, v_2)$ be the minimizing vector of $K()$, retaining  an implicit dependence on $n$ and $\lambda$.
We will be making frequent use of the partial derivatives of $K$:
\begin{align}
\frac{\partial K}{\partial w_1} &=   \frac{-9}{10 (1+ \exp(w_1+w_2) )} + \frac{1}{10 (1+ \exp(-w_1 + w_2) )} + \lambda w_1 \\
\frac{\partial K}{\partial w_2} &=   \frac{-9}{10 (1+ \exp(w_1+w_2) )} + \frac{-1}{10 (1+ \exp(-w_1 + w_2) )} + (n-1) \lambda w_2  .
\end{align}

It suffices to show that $0 \leq v_1 < v_2$ so that the first feature does not perturb the majority vote of the others.

To see  $0 \leq v_1$, notice that $\partial K / \partial w_1 \big|_{(0,w_2)}$ is negative for all $w_2$, including when $w_2 = v_2$.

To prove $v_1 < v_2$ we show the existence of a point $(a,a)$ such that
\begin{equation}
\label{e:partial.condition}
 \frac{\partial K}{\partial w_1} \Bigg|_{(a,a)} = -  \frac{\partial K}{\partial w_2} \Bigg|_{(a,a)} > 0,
\end{equation}
so that Lemma~\ref{l:halfspace} implies that the optimizing $(v_1,v_2)$ lies above the $w_1 = w_2$ diagonal.

\centerline{
\begin{tikzpicture}
  \draw [->] (0,0) -- node[left] {$w_2$}  (0,3);
  \draw [->] (0,0) -- node[below] {$w_1$} (3,0);
  \draw [dashed] (0,0) -- ( 2.5,  2.5);
  \node at (1.5,1.5)   [label=below:{$(a,a)$}]  {};
  \node at (1.5,1.5)   [circle, fill=blue, inner sep=1.5pt ]  {};   
  \draw [->,red] (1.5, 1.5) --   (0.7, 2.3) node[right]{$-\nabla K$} ;
\end{tikzpicture}
}

We have
\[
\frac{\partial K }{  \partial w_1}  \Big|_{(a,a)} =  \frac{-9}{10(1+\exp(2a))} + \frac{1}{20} + \lambda a
\]
which is increasing in $a$, negative when $a=0$ and goes to infinity with $a$.
It turns positive at some $a  < 1.5$ (exactly where depends on $\lambda$).

On the other hand,
\[
\frac{\partial K }{ \partial w_2 } \Big|_{(a,a)} =  \frac{-9}{10 (1+\exp(2a))} + \frac{-1}{20} + \lambda (n-1) a
\]
and is also increasing in $a$ and goes to infinity.
However, $\partial K / \partial w_2 \Big|_{(a,a)}$
is negative at $a=1.5$ whenever  $1.5 \lambda(n-1) \leq 1/20$, which is implied by the premise of the theorem.

Both partial derivatives are negative when $a=0$, continuously go to infinity with $a$, and $\partial K / \partial w_1 \Big|_{(a,a)}$ crosses zero first.
From the point where $\partial K / \partial w_1 \Big|_{(a,a)}$
crosses zero until $\partial K / \partial w_2 \Big|_{(a,a)}$ does,
the magnitude of $\partial K / \partial w_1 \Big|_{(a,a)}$
is increasing, starting at $0$, and the magnitude of $\partial K / \partial w_2 \Big|_{(a,a)}$
is decreasing until it reaches $0$.  When they meet, Equation~(\ref{e:partial.condition}) holds, completing the proof.

\section{Proof of Theorem~\ref{t:dropout.fails.large.n}}
\label{a:dropout.fails.large.n}
\begin{quote}
{\bf Theorem~\ref{t:dropout.fails.large.n}.}
If the dropout probability $q=1/2$ and the number of features is an even $n > 125$
then the weight vector $\bw^*(\PLtwo, q)$ optimizing the dropout criterion
has prediction error rate $\er_{\PLtwo} (\bw^*(\PLtwo, q)) \geq 1/10$.
\end{quote}
In this proof, we again abbreviate, using $P$ for $\PLtwo$.
  
The complicated form of the criterion optimized by dropout makes analyzing it difficult.
Here we make use of Jensen's inequality.
However, a straightforward application of it is fruitless, and a key step is to apply Jensen's inequality on just half the distribution resulting from dropout.

Similarly to before, let
\begin{equation}
\label{e:J.dropout.fails.large.n}
J(\bw) = \E_{(\bx,y) \sim P, \br}  ( \ell(y (\bw \cdot (\br \odot \bx))) ),
\end{equation}
and let $\bw^{\circledast}$ minimize $J$, so that $\bw^* = p \bw^{\circledast}$.

Again using symmetry and convexity, the last $n-1$ components of the optimizing $\bw^{\circledast}$ are equal,
so $\bw^{\circledast}$ is of the form $(w^{\circledast}_1, w^{\circledast}_2, w^{\circledast}_2, \ldots, w^{\circledast}_2)$.

\begin{lemma}
The minimizing $w^{\circledast}_1$ of ~(\ref{e:J.dropout.fails.large.n}) is positive.
\end{lemma}
\textbf{Proof:}
Let  $\widetilde{P,\br} $ be the marginal distribution of the last $n-1$ components after dropout and $\tilde{\bx}$ denote these last $n-1$ components of the dropped-out feature vector.
Then, recalling $y$ is always 1 in our distribution
(and $p$ is the probability that the first feature is \emph{not} dropped out),
\[
\frac{\partial J(w)}{\partial w_1} = \E_{(r_2,...,r_n)} \left( \frac{9 p}{10} \E_{\tilde{\bx} \sim {\widetilde{P, \br} }} ( \ell' (\bw \cdot (1, \tilde{\bx}) ) )
						- \frac{p}{10} \E_{\tilde{\bx} \sim {\widetilde{P, \br} }} ( \ell' (\bw \cdot (-1, \tilde{\bx}) ) ) \right)
\]
which is negative whenever $w_1=0$, since $\ell'()$ is negative and the two inner expectations become identical when $w_1=0$.
Therefore the optimizing $w^{\circledast}_1$  is positive.
\qed

To show that dropout fails, we want to show that $w^{\circledast}_1 > w^{\circledast}_2$, i.e.\ that $w^{\circledast}_1 \leq w^{\circledast}_2$ leads to a contradiction, so we begin to explore the consequences of $w^{\circledast}_1 \leq w^{\circledast}_2$.
\begin{lemma}
\label{l:w2.big.large.n}
If $q=1/2$ and $w^{\circledast}_1 \leq w^{\circledast}_2$ then $w^{\circledast}_2 > 4/9$.
\end{lemma}
\textbf{Proof:}
Assume to the contrary that $w^{\circledast}_1 \leq w^{\circledast}_2 \leq 4/9$.

Using Jensen's inequality,
\[
J(\bw^{\circledast})   \geq  \ell( \E_{(\bx,y) \sim P, \br} (y (\bw^{\circledast} \cdot \bx) ) )
\]
and the inner expectation is $8 w^{\circledast}_1 /  20 +  w^{\circledast}_2/2  \leq 9 w^{\circledast}_2 / 10$ as $w_1^{\circledast} \leq w_2^{\circledast}$.
Therefore, since $w^{\circledast}_2 \leq 4/9$,
\[
J(\bw^{\circledast}) \geq \ell(0.4) > 0.51.
\]
However,
\[
J(2.1, 0, 0, \ldots, 0) = \frac{\ln(2)}{2} + \frac{9 \ln(1+e^{-2.1} )}{20} + \frac{\ln(1+e^{2.1})}{20} < 0.51
\]
contradicting the optimality of $\bw^{\circledast}$.
\qed

\begin{lemma}
\label{l:binom.bound}
If $q=1/2$ and $w^{\circledast}_1 \leq w^{\circledast}_2$ then
$J(\bw^{\circledast}) \geq \E_{k \sim B(n, 1/2)} \ell(w^{\circledast}_2  (k - (n/2) + 1) )$
where $B(n,1/2)$ is the binomial distribution.
\end{lemma}
\textbf{Proof:}
Consider the modified distribution $P_1$ over $(\bx, y)$ examples where $y$ is always 1,
$x_2$, ..., $x_n$ are uniformly distributed over the the vectors with $n/2$ ones and $(n/2) - 1$  negative ones
(as in $P$), but $x_1$ is always one.
Since $0 < w^{\circledast}_1  \leq w^{\circledast}_2$ and the label $y=1$ under $P$ and $P_1$,
\begin{align*}
J(\bw^{\circledast}) & = \E_{(\bx,y) \sim P, \br}  ( \ell(\bw^{\circledast} \cdot  \bx) ) \\
	& >  \E_{(\bx,y) \sim P_1, \br}  ( \ell (\bw^{\circledast} \cdot \bx) ) \\
	& = \E_{(\bx,y) \sim P_1, \br}  \left( \ell \big(w^{\circledast}_2 (\mathbf{1} \cdot ( \bx \odot \br ) ) \big)  \right) \\
	& = \E_{(\bx,y) \sim P_1, \br}  \left( \ell \big(w^{\circledast}_2 (\bx \cdot \br) \big)  \right). \\
\end{align*}
Every $\bx$ in the support of  $P_1$ has exactly $(n/2)+1$ components that are 1, and the remaining $(n/2) -1$ components
are $-1$.
Call a component a \emph{success} if it is either $-1$ and dropped out or $1$ and not dropped out.
Now, $\bx \cdot \br$ is exactly $1 - (n/2) $ plus the number of successes.
Furthermore, the number of successes is distributed according to the binomial distribution $B(n,1/2)$.
Therefore
\[
\E_{(\bx,y) \sim P_1, \br}  ( w^{\circledast}_2 ( \bx \cdot \br ))  = \E_{k \sim B(n, 1/2)} (\ell(w^{\circledast}_2  (k - (n/2) + 1) ))
\]
giving the desired bound.
\qed

\begin{lemma}
\label{l:binom.big}
For even $n \geq 6$,
$ \E_{k \sim B(n, 1/2)} (\ell(w^{\circledast}_2  (k - (n/2) + 1) )) \geq \frac{1}{3}    \ell \left( w^{\circledast}_2  -   \frac{ w^{\circledast}_2 \sqrt{2n}} { 4 } \right)$.
\end{lemma}
\textbf{Proof:}
Let $\alpha = \sum_{i=0}^{n/2 -1} { n \choose i}$, so $\alpha$ is slightly less than $2^{n-1}$.
\begin{align*}
 \E_{k \sim B(n, 1/2)} (\ell(w^{\circledast}_2  (k - (n/2) + 1) ) )
 & = \frac{1}{2^n}  \sum_k {n \choose k}  \ell(w^{\circledast}_2 (k+ 1 - (n/2))) \\
 &  > \frac{\alpha}{2^n}  \sum_{k=0}^{n/2 -1}  \frac{1}{\alpha} {n \choose k}   \ell(w^{\circledast}_2 (1 + k - (n/2))) \\
 & >  \frac{\alpha}{2^n}    \ell \left( \sum_{k=0}^{n/2 - 1}  \frac{1}{\alpha} {n \choose k} w^{\circledast}_2 (1 + k  - (n/2)) \right)
\end{align*}
where the last step uses Jenson's inequality.
Continuing,
\begin{align*}
 \E_{k \sim B(n, 1/2)} (\ell(w^{\circledast}_2  (k - (n/2) + 1) ) )
 & >  \frac{\alpha}{2^n}    \ell \left( w^{\circledast}_2 +  \frac{w^{\circledast}_2}{\alpha}  \sum_{k=0}^{n/2 - 1}  {n \choose k} (k  - (n/2)) \right).
\end{align*}

Equation (5.18) of Concrete Mathematics~\cite{GKP89} and the bound
$
{n \choose n/2} \geq \frac{2^n}{ \sqrt{2n} }
$  
give
\[
\sum_{k=0}^{n/2 - 1}  {n \choose k} (k  - (n/2)) = \frac{- n } { 4} {n \choose n/2} \leq  \frac{-\sqrt{2n} \; 2^{n-1} }{4}.
\]

Therefore, recalling  that $\alpha< 2^{n-1}$ and noting $\alpha / 2^n > 1/3$ when $n \geq 6$,
\begin{align*}
 \E_{k \sim B(n, 1/2)} (\ell(w^{\circledast}_2  (k - (n/2) + 1) ) )
& >  \frac{\alpha}{2^n}    \ell \left( w^{\circledast}_2  -   \frac{w^{\circledast}_2}{\alpha}   \frac{ 2^{n-1}\sqrt{2n}} {4} \right) \\
& >  \frac{1}{3}    \ell \left( w^{\circledast}_2  -   \frac{ w^{\circledast}_2 \sqrt{2n}} { 4 } \right).
\end{align*}
\qed

We now have the necessary tools to prove Theorem~\ref{t:dropout.fails.large.n}.

\textbf{Proof: (of Theorem~\ref{t:dropout.fails.large.n}) }
If $w^{\circledast}_1 > w^{\circledast}_2$ then the first feature will dominate the majority vote of the others and the optimizing $\bw^{\circledast}$ has prediction error rate $1/10$ .
We now assume to the contrary that $w^{\circledast}_1 \leq w^{\circledast}_2$.
When $n >  125$ and $w^{\circledast}_2 \geq 4/9$ (from Lemma~\ref{l:w2.big.large.n})
we have
\[
w^{\circledast}_2  -   \frac{ w^{\circledast}_2 \sqrt{2n}} { 4 } \leq - 1.31
\]
and $\ell( w^{\circledast}_2  -   \frac{ w^{\circledast}_2 \sqrt{2n}} { 4 } )  > 1.54$.

Lemmas~\ref{l:binom.bound} and~\ref{l:binom.big} now imply that $J(\bw^{\circledast}) > 0.51$,
but (as in Lemma~\ref{l:w2.big.large.n})
$J(2.1,  0, 0, \ldots 0) < 0.51$, contradicting the optimality of $\bw^{\circledast}$.
\qed

Many of the approximations used to prove Theorem~\ref{t:dropout.fails.large.n}  are quite loose,
resulting in large values of $n$ being needed to obtain the contradiction.
For this class of distributions and $q=1/2$ we conjecture that optimizing the dropout criterion fails to produce the Bayes optimal hypothesis for
every even $n \geq 4$.

\section{Proof of Theorem~\ref{t:dropout.fails.n.eq.4}}
\label{a:dropout.fails.n.eq.4}
\begin{quote}
{\bf Theorem~\ref{t:dropout.fails.n.eq.4}.}
If dropout probability $q=1/2$ and the number of features is $n =4$
then the minimizer of the dropout criteria $\bw^*(\PLtwo, q)$ has 
has prediction error rate $\er_{\PLtwo} (\bw^*(\PLtwo, q)) \geq 1/10$.
\end{quote}

In this proof, let us also refer to $\PLtwo$ as just $P$ and let $\bw^{\circledast}$ be the minimizer of 
(\ref{e:J.dropout.fails.large.n}).

As before, the optimizing $\bw^{\circledast}$ has the form $(w^{\circledast}_1, w^{\circledast}_2, w^{\circledast}_2, w^{\circledast}_2)$ by symmetry and convexity.
Recalling that the label $y$ is always $1$ under distribution $P$, we can use the equivalent criterion
\[
K(w_1, w_2) =
\E_{(\bx,y) \sim P, \br}  ( \ell(y (\bw \cdot \bx)) ) = \E_{(\bx,y) \sim P, \br}  \left( \ell \left(w_1 x_1 r_1 + w_2 \sum_{i=2}^4 x_i r_i  \right ) \right) .
\]
This expectation can be written with 12 terms, one for each pairing of the three possible $x_1 r_1$ values with the   four possible
$\sum_{i=2}^4 x_i r_i \in \set{-1, 0, 1, 2}$ values (see Table~\ref{t:probs}).

\begin{table}
\caption{Probabilities of $x_1 r_1$ and $\sum_{i=2}^4 x_i r_i $ values assuming dropout probability $q=1/2$.}
\label{t:probs}

\smallskip


\centerline{
\begin{tabular}{|c c | c c |}
\hline
$x_1 r_1$ & probability & $\sum_{i=2}^4 x_i r_i $ & probability \\ \hline
1	& 9/20	&  	2	& 	1/8 \\
0	& 1/2		&  	1	& 	3/8 \\
-1	& 1/20	&  	0	& 	3/8 \\
	&		&	-1	&	1/8 \\ \hline
\end{tabular}
}
\end{table}

Taking them in order, we have
\begin{align*}
K(w_1, w_2) = &
   \frac{9}{160} \ell \left(w_1 + 2 w_2 \right )
   + \frac{27}{160} \ell \left(w_1 + w_2 \right )
   + \frac{27}{160} \ell \left(w_1 \right )
   + \frac{9}{160} \ell \left(w_1 - w_2 \right ) \\
  &    + \frac{10}{160} \ell \left(2 w_2 \right )
       + \frac{30}{160} \ell \left(w_2 \right )
       + \frac{30}{160} \ell \left(0 \right )
       + \frac{10}{160} \ell \left(w_2 \right ) \\
  &  + \frac{1}{160} \ell \left(-w_1 + 2 w_2 \right )
   + \frac{3}{160} \ell \left(-w_1 + w_2 \right )
   + \frac{3}{160} \ell \left(-w_1 \right )
   + \frac{1}{160} \ell \left(-w_1 - w_2 \right ).
\end{align*}

So, when $p=q=1/2$, the derivatives are:
\begin{align*}
\frac{\partial K}{\partial w_1} = & \Bigg(
	     \frac{-9}{1 + \exp(w_1+2 w_2)}
	+  \frac{-27}{1 + \exp(w_1+ w_2)}
	+  \frac{-27}{1 + \exp(w_1)} +  \frac{-9}{1 + \exp(w_1 - w_2)}  \\
& \quad	+    \frac{1}{1 + \exp(-w_1+2 w_2)}
        +  \frac{3}{1 + \exp(-w_1+ w_2)}
        +  \frac{3}{1 + \exp(-w_1)}
	+  \frac{1}{1 + \exp(-w_1 - w_2)}
	 \Bigg) \Big/ 160 ,  \\
\frac{\partial K}{\partial w_2} = & \Bigg(
	     \frac{-18}{1 + \exp(w_1+2 w_2)}
	+  \frac{-27}{1 + \exp(w_1+ w_2)}
	+  \frac{9}{1 + \exp(w_1-w_2)}  \\
& \quad	+  \frac{-20}{1 + \exp(2 w_2)}
	+  \frac{-30}{1 + \exp( w_2)}
	+  \frac{10}{1 + \exp(- w_2)} \\
& \quad	+  \frac{-2}{1 + \exp(-w_1 + 2 w_2)}
	+    \frac{-3}{1 + \exp(-w_1+ w_2)}
	+  \frac{1}{1 + \exp(-w_1 - w_2)}
\Bigg) \Big / 160 .
\end{align*}

If $w^{\circledast}_1 > w^{\circledast}_2$, then dropout will have prediction error rate 1/10 as $w^{\circledast}_1$ will dominate the vote of the other three components.
We show that $w^{\circledast}_1 >  w^{\circledast}_2$ by proving that there is a point $(a,a)$ in weight space such that the
gradient at $(a,a)$ is of the form $(-c, c)$ for some $c > 0$ (see Figure~\ref{f:grad.w1.bigger}).

\begin{figure}
\centerline{
\begin{tikzpicture}
  \draw [->] (0,0) -- node[left] {$w_2$}  (0,3);
  \draw [->] (0,0) -- node[below] {$w_1$} (3,0);
  \draw [dashed] (0,0) -- ( 2.5,  2.5);
  \node at (1.5,1.5)   [label=below:{ \small $(a,a)$}]  {};
  \node at (1.5,1.5)   [circle, fill=blue, inner sep=1.5pt ]  {};   
  \draw [->,blue] (1.5, 1.5) --   (0.7, 2.3) node[above]{\small \qquad $\nabla K  =  (-c,c)$} ;
  \draw [->,red] (1.5, 1.5) --   ( 2.3,  0.7) node[right]{ \small $-\nabla K  =  (c,-c)$} ;
\end{tikzpicture}
}
\caption{If $\nabla K$ at some $(a,a)$ is $(-c, c)$ for some $c>0$ then $w^{\circledast}_1 > w^{\circledast}_2$.}
\label{f:grad.w1.bigger}
\end{figure}

The derivatives when evaluated at $(a,a)$ are:
\begin{align*}
\frac{\partial K}{\partial w_1} \Bigg |_{(a,a)}= & \Bigg(
	\frac{-9}{1+\exp(3a)}
	+ \frac{-27}{1+\exp(2a)}
	+ \frac{-26}{1+\exp(a)}
	- 3
	+ \frac{3}{1+\exp(-a)}
	+ \frac{1}{1+\exp(-2a)}
	\Bigg) \Big / 160 \\
\frac{\partial K}{\partial w_2} \Bigg |_{(a,a)}= & \Bigg(
	\frac{-18}{1+\exp(3a)}
	+ \frac{-47}{1+\exp(2a)}
	+ \frac{-32}{1+\exp(a)}
	+3
	+ \frac{10}{1+\exp(-a)}
	+ \frac{1}{1+\exp(-2a)}
	\Bigg) \Big / 160 .
\end{align*}
Note that both of these derivatives are increasing in $a$,  positive for large $a$, and negative when $a=0$.
At $a=2 \ln(2)$, derivative $\partial K / \partial w_1 \big |_{(a,a)}$ is still negative, while
$\partial K / \partial w_1 \big |_{(a,a)}$ has turned positive,
so $\partial K / \partial w_1 \big |_{(a,a)}$ crosses 0 first.
The continuity of the partial derivatives now implies the existence of an $(a,a)$ where $\nabla K$ has the form $(-c, c)$,
completing the proof.
\qed

\section{Proof of Theorem~\ref{t:dropout.succeeds.n>1}}
\label{a:dropout.succeeds.n>1}
\begin{quote}
{\bf Theorem~\ref{t:dropout.succeeds.n>1}.}
If $q = 1/2$, $n \geq 100$, $\alpha > 0$,
$\beta = 1/(10 \sqrt{n-1})$, and
$\eta \leq \frac{1}{2 + \exp(54 \sqrt{n})}$,
then
$
\er_{\Pdrop} (\bw^*(\Pdrop, q)) = \eta.
$
\end{quote}

For this subsection, let $P = \Pdrop$ and define the scaled dropout criterion
\[
J(\bw) = \E_{(\bx,y) \sim P, \br}
                  ( \ell(y \bw \cdot (\br \odot \bx)) ),
\]
where, as earlier, the components of $\br$ are independent
samples from a Bernoulli distribution with parameter $p=1-q=1/2>0$.
Let $\bw^{\circledast}$ be the minimizer of $J$, so that
$\bw^* = p \bw^{\circledast}$.

Note that, by symmetry, the contribution to $J$ from the cases where
$y$ is $-1$ and $1$ respectively are the same, so the value of $J$ is
not affected if we clamp $y$ at $1$.  Let us use this form to express
$J$, and let $D$ be the marginal distribution of feature vector $\bx$ conditioned on the label $y=1$.

Let $B = \{ 2,...,n \}$.  By symmetry, $w^{\circledast}_i$ is identical for all $i
\in B$ so $\bw^{\circledast}$ is the minimum of $J$ over weight vectors satisfying
this constraint.  Let $K(w_1,w_2) = J(w_1,w_2,...,w_2)$; note that $w_1^{\circledast},w_2^{\circledast}$ minimizes
$K$ defined by
\[
K(w_1, w_2) =
    \E_{\bx \sim D, \br} (\ell(w_1 r_1 x_1  + w_2 \sum_{i \in B} r_i x_i )).
\]

To prove Theorem~\ref{t:dropout.succeeds.n>1}, it suffices to show
that
\begin{equation}
\label{e:suffices}
w_1^{\circledast} > (n-1) w_2^{\circledast}/\alpha > 0,
\end{equation}
since when~(\ref{e:suffices}) holds, $\bw^{\circledast}$
always outputs $x_1$.

We have
\begin{align}
\label{e:partial.w1.n>1}
\frac{\partial K}{\partial w_1}
& =  \frac{1}{2} \E_{\bx \sim D, \br} \left(\frac{-x_1}{1 + \exp(w_1 x_1 + w_2 \sum_{i \in B} r_i x_i )}\right) \\
\label{e:partial.wB.n>1}
\frac{\partial K}{\partial w_2}
& = \E_{\bx \sim D, \br} \left(\frac{-\sum_{i \in B} r_i x_i}{1 + \exp(w_1 r_1 x_1 + w_2 \sum_{i \in B} r_i x_i )}\right).
\end{align}
(Note that, in (\ref{e:partial.w1.n>1}), we have marginalized out $r_1$.)

\begin{lemma}
\label{l:wB>0.n>1}
$w^{\circledast}_2 > 0$.
\end{lemma}
As before, it suffices to show that there is a point
$(a_1,0)$ where both $\frac{\partial K}{\partial w_2} \big|_{(a_1,0)} <0$ and
$\frac{\partial K}{\partial w_1} \big|_{(a_1,0)} = 0$.   From
equation (\ref{e:partial.wB.n>1}),
\[
\frac{\partial K}{\partial w_2} \big|_{(a_1,0)} =  \E_{\bx \sim D, \br} \left(\frac{-\sum_{i \in B} r_i x_i}{1 + \exp(a_1 r_1 x_1)}\right) < 0
\]
for all real $a_1$.

Now, evaluating (\ref{e:partial.w1.n>1}), dividing into cases based on $x_1$, we get
\[
\frac{\partial K}{\partial w_1} \big|_{(a_1,0)}
 = (\eta/2) \left( \frac{\alpha}{1 + \exp(- \alpha a_1)} \right)
  + ((1 - \eta)/2) \left(\frac{-\alpha}{1 + \exp(\alpha a_1)}\right).
\]
This approaches $-\alpha ((1 - \eta)/2)$ as $a_1$ approaches $-\infty$, and it approaches $\alpha \eta/2$ as $a_1$ approaches
$\infty$.  Since it is a continuous function of $a_1$, there must be a value of $a_1$ such that
$\frac{\partial K}{\partial w_1} \big|_{(a_1,0)} = 0$.  Putting this together with $\frac{\partial K}{\partial w_2} \big|_{(a_1,0)} <0$
completes the proof.  \qed

To show the sufficient inequalities~(\ref{e:suffices}), it will be useful to prove an upper
bound on $w_2^{\circledast}$.  (This upper bound will make it easier to show, informally, that $w_1^{\circledast}$ is needed.)  In
order to bound the size of $w_2^{\circledast}$, we will prove a lower bound on $K$
in terms of $w_2$.  For this, we want to show that, if $w_2$ is too
large, then the algorithm will pay too much when it makes large-margin
errors.  For {\em this}, we need a lower bound on the probability of a
large-margin error.  For this, we can adapt an analysis that provided
a lower bound on the probability of an error from \cite{HL12}.

To simplify the proof, we will first provide a lower bound on the
dropout risk in terms of the risk without dropout.  We will actually
prove something somewhat more general, for possible future reference.
\begin{lemma}
\label{l:remove.dropout}
Let $\br$ and $\bx$ be independent, $\R^N$-valued random variables;
let $\phi$ be convex function of a scalar real variable.  Then
\[
\E_{\br, \bx} \left(\phi\left(\sum_i x_i r_i\right)\right)
  \geq
    \E_{\bx} \left(\phi\left(\sum_i x_i \E_{\br}(r_i)\right)\right).
\]
\end{lemma}
{\bf Proof:}  Since $\bx$ and $\br$ are independent,
\begin{align*}
& \E_{\br, \bx} (\phi(\sum_i x_i r_i)) \\
& = \E_{\bx} (\E_{\br} (\phi(\sum_i x_i r_i))) \\
& \geq \E_{\bx} (\phi(\E_{\br} (\sum_i x_i r_i)))
    \hspace{0.1in} \mbox{(by Jensen's Inequality)} \\
& = \E_{\bx} (\phi(\sum_i x_i \E_{\br} (r_i))),
\end{align*}
completing the proof. \qed

Now, it is enough to lower bound the probability of a large-margin
error with respect to the original distribution.  Recall $B = \{ 2, \ldots, n \}$.

\begin{lemma}
\label{l:slud}
\ \
\( \displaystyle
\Pr\left( \frac{1}{n-1} \sum_{i \in B} x_i < -2 \beta \right) \geq \frac{3}{10}.
\)
\end{lemma}
{\bf Proof:}  If $Z$ is a standard normal random variable
and $R$ is a binomial $(\ell , p )$ random variable with $p\leq  1/2$,
then for $\ell(1-p) \leq j \leq \ell p$, Slud's
inequality \cite{Slu77} (see also Lemma~23 of \cite{HL12})
%
gives
\begin{equation}
\label{e:HL}
\Pr(R \geq j) \geq \Pr \left( Z \geq \frac{j- \ell p}{\sqrt{\ell p (1-p)}} \right).
\end{equation}

Now, we have
\begin{align*}
\Pr \left( \frac{1}{n-1} \sum_{i \in B} x_i < -2 \beta \right)
	&= \Pr \left(  \sum_{i \in B} x_i/2  < - (n-1) \beta \right)  \\
	&= \Pr \left(  \sum_{i \in B} (x_i + 1)/2 < (n-1)/2-(n-1) \beta \right)  \\
	&= \Pr \left(\sum_{i \in B} z_i < (n-1)(1/2 - \beta)  \right)
\end{align*}
where the $z_i$'s are independent $\set{0,1}$-valued variables with $\Pr(z_i=1) = 1/2 + \beta$.
Let $\bar z_i$ be $1-z_i$, so $\sum_{i \in B} \bar z_i$ is a Binomial $(n-1, 1/2 - \beta)$ random variable.  Furthermore,
\begin{align*}
\Pr \left(\sum_{i \in B} z_i < (n-1)(1/2 - \beta)  \right)
	 & = \Pr \left( \sum_{i \in B} \bar z_i > (n-1) - (n-1)(1/2 - \beta)  \right)  \\
	&  = \Pr \left( \sum_{i \in B} \bar z_i > (n-1) (1/2 + \beta)  \right) .
\end{align*}
Using (\ref{e:HL}) with  $j=(n-1)(1/2+\beta)$, $\ell = (n-1)$, and $p = 1/2 - \beta$ gives:
\begin{align*}
\Pr \left( \sum_{i \in B} \bar z_i > (n-1) (1/2 + \beta)  \right)
	 & \geq \Pr \left( Z \geq \frac{(n-1)(1/2 + \beta) - (n-1) (1/2 - \beta)}{\sqrt{(n-1) (1/4 - \beta^2)}} \right) \\
	 & = \Pr \left( Z \geq \frac{2(n-1)\beta}{\sqrt{(n-1) (1/4 - \beta^2)}} \right).
\end{align*}

Since $\beta = 1/(10 \sqrt{n})$ and $n \geq 100$, this implies
\[
\Pr\left( \frac{1}{n-1} \sum_{i \in B} x_i < -2 \beta \right)
  \geq
   \Pr\left( Z \geq 1/2
             \right).
\]
Since the density of $Z$ is always at most $1/\sqrt{2 \pi}$, we have
\[
\Pr\left( \frac{1}{n-1} \sum_{i \in B} x_i < -2 \beta \right)
 \geq \Pr(Z \geq 0) - \Pr(Z \in (0,1/2))
 > \frac{1}{2} - \frac{1}{2 \sqrt{2 \pi}} > 3/10 ,
\]
completing the proof. \qed

Now we are ready for the lower bound on the dropout risk in terms of $w_2$.
\begin{lemma}
\label{l:lower.K.w2}
For all $w_1$,
\[
K(w_1,w_2) > \frac{w_2 \sqrt{n-1}}{67}.
\]
\end{lemma}
{\bf Proof:}  Considering only the case in which $x_1$ is dropped out (i.e. $r_1 = 0$), we have
\[
K(w_1,w_2) \geq \frac{1}{2} \E\left( \ell\left(w_2 \sum_{i} r_i x_i\right) \right).
\]
Applying Lemma~\ref{l:remove.dropout}, we get
\[
K(w_1,w_2) \geq \frac{1}{2} \E\left( \ell\left((w_2/2) \sum_{i \in B} x_i \right) \right).
\]
Since $\ell$ is non-increasing and non-negative, we have
\[
K(w_1,w_2) \geq \frac{1}{2} \ell( - w_2 \beta (n-1))
                           \Pr\left( \frac{1}{n-1}
                         \sum_{i \in B} x_i < -2 \beta \right),
\]
and applying Lemma~\ref{l:slud} gives
\[
K(w_1,w_2) \geq \frac{3 \ell( - w_2 \beta (n-1))}{20}.
\]
Since $\ell(z) > -z$, we have
\[
K(w_1,w_2) \geq \frac{3 w_2 \beta (n-1)}{20}
\]
and, using $\beta = \frac{1}{10 \sqrt{n-1}}$,
we get
\[
K(w_1,w_2) \geq \frac{ 3 w_2 \sqrt{n-1}}{200},
\]
completing the proof. \qed

\begin{lemma}
\label{l:w2.small.n>1}
$w^{\circledast}_2 < \frac{27}{\sqrt{n-1}}$.
\end{lemma}
{\bf Proof}:  Note that
\[
K(w, 0)
  = \ell(0)/2 + (1/2) ( \eta \ell(- \alpha w) + (1 - \eta) \ell(\alpha w)),
\]
is increasing in $\eta$ so that
\begin{equation}
\label{r:benchmark}
K(w^{\circledast}_1, w^{\circledast}_2) \leq K(5 / \alpha, 0) <
\ell(0)/2 + 1/35
\end{equation}
since $\eta < 1/100$.

On the other hand, Lemma~\ref{l:lower.K.w2} gives
\[
K(w^{\circledast}_1, w^{\circledast}_2) > \frac{w_2 \sqrt{n-1}}{67}.
\]
Solving for $w_2^{\circledast}$ completes the proof.
\qed

\begin{lemma}
\label{l:w1.partial.n>1}
For all $0 < u < \frac{27}{\sqrt{n-1}}$, we have
\[
\frac{\partial K}{\partial w_1} \big|_{((n-1) u/\alpha,u)} < 0.
\]
\end{lemma}
{\bf Proof:}  From (\ref{e:partial.w1.n>1}), we have
\begin{align*}
& 2 \frac{\partial K}{\partial w_1}\big|_{(n u/\alpha,u)} \\
& = \E_{\bx \sim D, \br} \left(\frac{-x_1}{1 + \exp((n-1) u x_1/\alpha + u \sum_{i \in B} r_i x_i )}\right) \\
& = \eta \E_{\bx \sim D, \br} \left(\frac{\alpha}{1 + \exp(-(n-1) u  + u \sum_{i \in B} r_i x_i )}\right)
       + (1 - \eta) \E_{\bx \sim D, \br} \left(\frac{- \alpha}{1 + \exp((n-1) u + u \sum_{i \in B} r_i x_i )}\right) \\
& < \eta \alpha + (1 - \eta) \E_{\bx \sim D, \br} \left(\frac{- \alpha}{1 + \exp((n-1) u + u \sum_{i \in B} r_i x_i )}\right) \\
& < \alpha \left(\eta + \frac{-(1 - \eta)}{1 + \exp(2 (n-1) u)}\right) \hspace{0.1in}\mbox{(since $\sum_{i \in B} r_i x_i \leq n-1$)} \\
& < \alpha \left(\eta + \frac{-(1 - \eta)}{1 + \exp(54 \sqrt{n-1}}\right) \hspace{0.1in}\mbox{(since $u < 27/\sqrt{n-1}$)} \\
& <  0
\end{align*}
since $\eta \leq 1/(2+ \exp(54 \sqrt{n}))$, completing the proof.
\qed

Recall that, to prove Theorem~\ref{t:dropout.succeeds.n>1}, since we already showed $w_2^{\circledast} > 0$, all we needed was to show that $\alpha w_1^{\circledast} > (n-1) w_2^{\circledast}$.
We do this next.
\begin{lemma}
$\alpha w_1^{\circledast} > (n-1) w_2^{\circledast}$.
\end{lemma}
{\bf Proof}:  Let $\bg$ be the gradient of $J$ evaluated at $\bu = ((n-1) w^{\circledast}_2/\alpha, w^{\circledast}_2)$.
Lemmas~\ref{l:w2.small.n>1} and~\ref{l:w1.partial.n>1}
implies that $\bg \neq (0,0)$.  By convexity
\[
\bw^{\circledast} \cdot \bg < \bu \cdot \bg
\]
which, since $u_2 = w^{\circledast}_2$, implies
\[
w_1^{\circledast}  g_1 < (n-1) w^{\circledast}_2 g_1/\alpha.
\]
Since, by Lemmas~\ref{l:w2.small.n>1} and~\ref{l:w1.partial.n>1}, $g_1 < 0$,
\[
w_1^{\circledast} > (n-1) w^{\circledast}_2/\alpha
\]
completing the proof. \qed

\section{Proof of Theorem~\ref{t:elltwo.fails.n>1}}
\label{a:elltwo.fails.n>1}
\begin{quote}
{\bf Theorem~\ref{t:elltwo.fails.n>1}.}
If $\beta = 1/(10 \sqrt{n-1})$,
$\lambda = \frac{1}{30n}$,
$\alpha < \beta \lambda$,
and $n$ is a large enough even number, then for any $\eta \in [0,1]$,
$
\er_{\Pdrop} (\bv(\Pdrop, \lambda)) \geq 3/10.
$
\end{quote}

In this proof, let us also abbreviate $\Pdrop$ with $P$ and use $J$ to denote the $L_2$ regularized criterion in Equation~(\ref{e:L2.optimizer}) specialized for distribution this $P$.

As before, the contribution to the $L_2$ criteron from the cases where
$y$ is $-1$ and $1$ respectively are the same, so the value of the criterion is
not affected if we clamp $y$ at $1$.  
Furthermore, we leave the dependency on $\lambda$ implicit and
(since the source is fixed) use the more succinct $\bv$ for $\bv(P, \lambda)$.

Also, if, as before, we let $B = \{ 2,...,n \}$, then by symmetry,
$v_i$ is identical for all $i \in B$ so $\bv$ is the minimum of $J$
over weight vectors satisfying this constraint.  Let $K(w_1,w_2) =
J(w_1,w_2,...,w_2)$ so that $(v_1,v_2)$ minimizes $K$.
Recall that $D$ is the marginal distribution of $\bx$ under $P$ conditioned on $y=1$.
\[
K(w_1, w_2) =
    \E_{\bx \sim D} \left(\ell\left(w_1 x_1  + w_2 \sum_{i \in B} x_i \right)\right)
    + \frac{\lambda}{2} (w_1^2 + (n-1) w_2^2).
\]

Lemma~\ref{l:slud}, together with the fact that $|x_1| = \alpha$,
implies that,
\begin{equation}
\label{e:sufficient.elltwo.fails.n>1}
\alpha v_1 < 2 \beta (n-1) v_2
\end{equation}
suffices to prove Theorem~\ref{t:elltwo.fails.n>1}, so we set this as
our subtask.

We have
\begin{align}
\label{e:partial.w1.elltwo.n>1}
\frac{\partial K}{\partial w_1}
& =  \E_{\bx \sim D} \left(\frac{-x_1}{1 + \exp(w_1 x_1 + w_2 \sum_{i \in B} x_i )} \right)
         + \lambda w_1 \\
\label{e:partial.w2.elltwo.n>1}
\frac{\partial K}{\partial w_2}
& = \E_{\bx \sim D} \left(\frac{-\sum_{i \in B} x_i}{1 + \exp(w_1 x_1 + w_2 \sum_{i \in B} x_i )}\right)
         + \lambda (n-1) w_2.
\end{align}

First, we need a rough bound on $v_1$.
\begin{lemma}
\label{l:v1.rough}
$|v_1| \leq \frac{\alpha}{\lambda} < \beta$.
\end{lemma}
{\bf Proof:}
The second inequality follows from the constraint on $\alpha$.
From (\ref{e:partial.w1.elltwo.n>1}), we get
\[
|v_1| \leq \frac{1}{\lambda}
\E_{\bx \sim D} \left(\left| \frac{x_1}{1 + \exp(v_1 x_1 + v_2 \sum_{i \in B} x_i )}
              \right|\right)
\]
and the facts $|x_1| \leq \alpha$ and
$0 < \frac{1}{1 + \exp(v_1 x_1 + v_2 \sum_{i \in B} x_i )} \leq 1$
then imply $|v_1| \leq \alpha/\lambda$.
\qed


\begin{lemma}
\label{l:binomial.hump.low}
For large enough $n$,
\[
\Pr \left( \sum_{i \in B} x_i \in [\beta(n-1) , 3 \beta (n-1)] \right)
    \geq \frac{1}{13}.
\]
\end{lemma}
{\bf Proof:}  Let $\Phi(z) = \Pr(Z \leq z)$ for a standard normal
random variable $Z$ and let $S = \sum_{i \in B} x_i$.
Note that $\E(x_i) = 2 \beta$, $\mathbf{var}(x_i) = 1-4\beta^2$, and the third moment
$\E( | x_i  -\E(x_i) |^3) = 1- 16\beta^4$.
The Berry-Esseen inequality (see Theorem 11.1 of \cite{Das08}) relates binomial distributions to the normal distribution using these moments, and
directly implies that
\[
\sup_z \left|
 \Pr\left(\frac{S}{n-1} - 2 \beta
        \leq \sqrt{\frac{1 - 4 \beta^2}{n-1}} \times z
      \right)
   - \Phi(z) \right| \leq \frac{C (1-16 \beta^4)}{(1-4 \beta^2)^{3/2}\sqrt{n-1}} < \frac{1}{\sqrt{n-1}}
\]
where the last inequality follows from the facts  that the Berry-Esseen global constant $C \leq 0.8$ and $\beta < 1/10$.

Using the change of variable $s = \sqrt{(1-4\beta^2)(n-1)} \; z + 2\beta (n-1)$ this can be restated:
\[
\sup_s \left|
 \Pr\left(S \leq s \right)
   - \Phi\left(\frac{s - 2 \beta (n-1)}
                     {\sqrt{(1 - 4 \beta^2) (n-1)}}\right)
        \right| \leq \frac{1}{\sqrt{n-1}},
\]
so
\begin{align*}
& \Pr(S \in [ \beta (n-1), 3 \beta (n-1) ])  \\
& \geq
    \Pr_{z \in N(0,1)} \left(z \in \left[ -\beta \sqrt{ \frac{n-1}
                                                   {1 - 4 \beta^2}},
                                        \beta \sqrt{ \frac{n-1}
                                                   {1 - 4 \beta^2}}
                                          \right] \right)
                  - \frac{2}{\sqrt{n-1}} \\
& \geq
    \Pr_{z \in N(0,1)} \left(z \in \left[ \frac{-1}{10},
                                 \frac{1}{10} \right] \right)
                  - \frac{2}{\sqrt{n-1}} \\
& \geq \frac{1}{13},
\end{align*}
for large enough $n$. \qed

Recent work shows that the Berry-Esseen constant $C$ is less then 1/2, this allows us to replace the $2\sqrt{n-1}$
with $1/\sqrt{n-1}$, but it still requires $n$ on the order of 150,000 to get the 1/13 bound.
Reducing the bound to 1/50 would make $n$ as small as 300 sufficient.

Next, we need a rough bound on $v_2$.
\begin{lemma}
\label{l:v2.rough}
$v_2 \geq \frac{1}{n-1}$.
\end{lemma}
{\bf Proof:} From (\ref{e:partial.w2.elltwo.n>1}), we have
\[
v_2 = \frac{1}{\lambda (n-1)}
       \E_{\bx \sim D} \left(\frac{\sum_{i \in B} x_i}
                          {1 + \exp(v_1 x_1 + v_2 \sum_{i \in B} x_i )}\right).
\]
If we denote $\sum_{i \in B} x_i$ by $S$, then
\[
v_2 = \frac{1}{\lambda (n-1)}
       \E_{\bx \sim D} \left(\frac{S}
                          {1 + \exp(v_1 x_1 + v_2 S)}\right).
\]
Since, for all odd\footnote{$S$ is the sum of an odd number of $\pm1$'s, and thus cannot be even.} $s > 0$
\[
\frac{\Pr(S = s)}{\Pr(S = -s)}
    = \left( \frac{1 + 2 \beta}{1 - 2 \beta} \right)^s
\]
so  $\Pr(S=-s) = \Pr(S=s)  \left( \frac{1 - 2 \beta}{1 + 2 \beta} \right)^{s} $.
Analyzing the contributions of $s$ and $-s$ together
we have
\begin{align*}
v_2 \lambda (n-1) =
 \sum_{s=1}^{n-1}
  \Pr(S = s) \Big( & (1 - \eta) \frac{s}
                                      {1 + \exp(v_1 \alpha + v_2 s)}
                      + \eta \frac{s}
                                  {1 + \exp(-v_1 \alpha + v_2 s)} \\
                 &     +\left( (1 - \eta)
                            \frac{-s}
                                 {1 + \exp(v_1 \alpha - v_2 s)}
                      + \eta \frac{-s}
                                 {1 + \exp(-v_1 \alpha - v_2 s)} \right)
                     \left( \frac{1 - 2 \beta}{1 + 2 \beta} \right)^s \Big).
\end{align*}
Recalling that $|v_1| \leq \alpha / \lambda$ (Lemma~\ref{l:v1.rough}), and using the
minimizing value in this range for each term gives
\begin{align*}
v_2 \lambda (n-1) & \geq
 \sum_{s=1}^{n-1}
  \Pr(S = s) \left( \frac{s}
                          {1 + \exp(\alpha^2/\lambda + v_2 s)}
                     +\left( \frac{-s}
                                 {1 + \exp(-\alpha^2/\lambda - v_2 s)}
                         \right)
                     \left( \frac{1 - 2 \beta}{1 + 2 \beta} \right)^s \right)
   \\
& =
 \sum_{s=1}^{n-1}
  \Pr(S = s) s \left( \frac{1 - \exp(\alpha^2/\lambda + v_2 s)
                           \left( \frac{1 - 2 \beta}{1 + 2 \beta} \right)^s
                           }
                          {1 + \exp(\alpha^2/\lambda + v_2 s)}
                          \right)
   \\
& \geq
 \sum_{s=1}^{n-1}
  \Pr(S = s) s \left( \frac{1 - \exp(\alpha^2/\lambda + v_2 s - 4 \beta s)}
                          {1 + \exp(\alpha^2/\lambda + v_2 s)}
                          \right).
\end{align*}
Assume for contradiction that  $v_2 < 1/(n-1)$.  Then,
\begin{align*}
v_2 \lambda (n-1) & \geq
 \sum_{s=1}^{n-1}
  \Pr(S = s) s \left( \frac{1 - \exp(\alpha^2/\lambda + s/(n-1) - 4 \beta s)}
                          {1 + \exp(\alpha^2/\lambda + s/(n-1))}
                          \right) \\
  & \geq
 \sum_{s=1}^{n-1}
  \Pr(S = s) s \left( \frac{1 - \exp(s/(n-1) - 3 \beta s)}
                          {1 + \exp(\beta^2  \lambda + s/(n-1))}
                          \right)
      \hspace{0.1in} \mbox{(since $\alpha \leq \beta \lambda$)}
   \\
  & \geq
 \sum_{s=1}^{n-1}
  \Pr(S = s) s \left( \frac{1 - \exp(- 2 \beta s)}
                          {1 + \exp(\beta^2  \lambda + s/(n-1))}
                          \right)
      \hspace{0.1in} \mbox{(for large enough $n$)}
   \\
  & \geq
 \sum_{s \in [\beta (n-1) , 3 \beta (n-1) ]}
  \Pr(S = s) s \left( \frac{1 - \exp(- 2 \beta s)}
                          {1 + \exp(\beta^2  \lambda + s/(n-1))}
                          \right),
\end{align*}
since each term is positive.  Taking the worst-case among
$[\beta (n-1) , 3 \beta (n-1) ]$ for each instance of $s$, and
applying Lemma~\ref{l:binomial.hump.low},
we get
\begin{align}
v_2 & \geq \frac{1}{\lambda (n-1)} \times \frac{1}{13}
                  \times \beta (n-1)
               \left( \frac{1 - \exp(- 2 \beta^2 (n-1))}
                            {1 + \exp(\beta^2 \lambda + 3 \beta)}
                              \right) \nonumber \\
    & =  \frac{30 \sqrt{n-1}}{130}      \label{e:bound.rhs}
               \left( \frac{1 - \exp(- 1/50 )}
                            {1 + \exp(3/(10 \sqrt{n-1})
                            		+ 1/(3000 n(n-1)) )}
                              \right).
\end{align}
Thus $v_2 = \Omega(\sqrt{n-1})$, which, for large enough $n$,
contradicts our assumption that $v_2 < 1/(n-1)$, completing the proof.
\qed

Not that even with the many approximations made, Inequality~(\ref{e:bound.rhs}) gives the desired contradiction
at $n=60$.
Even when the weaker bound of 1/50 discussed following Lemma~\ref{l:binomial.hump.low} is used, $n=145$ still suffices to give the desired contradiction.

\medskip

Now we're ready to put everything together.

\medskip

\noindent
{\bf Proof (of Theorem~\ref{t:elltwo.fails.n>1}):}  Recall that, by
Lemma~\ref{l:slud}, if $v_1 < 2 \beta (n-1) v_2$, then
$\er_{P} (\bv(P, \lambda)) \geq 3/10.$

Lemma~\ref{l:v1.rough} gives $v_1 < \beta$.
Lemma~\ref{l:v2.rough} implies $(n-1) v_2 \geq 1$.
Therefore $v_1 < \beta (n-1) v_2$, completing the proof.  \qed

Using the 1/50 version of Lemma~\ref{l:binomial.hump.low} leads to a proof of the theorem for all even $n\geq 300$.

\end{document}